\documentclass[11pt]{article}

\usepackage[a4paper,margin=1in]{geometry}
\usepackage{amsmath,amssymb}
\usepackage{graphicx}
\usepackage{booktabs}
\usepackage{multirow}
\usepackage{array}
\usepackage{cite}
\usepackage[hidelinks]{hyperref}
\usepackage{url}
\usepackage{authblk}
\usepackage{graphicx}
\usepackage{pifont}   

\newcommand{\cmark}{\ding{51}}
\newcommand{\xmark}{\ding{55}}

\newcommand{\eg}{e.g.,}
\newcommand{\ie}{i.e.,}
\newcommand{\etal}{\textit{et~al.}}

\title{PaGNet: A Panel-Aware GBDT--Neural Network for
Multi-Target Corporate Tax Avoidance Proxy Forecasting}

\author[1]{Wonho Song}
\author[1]{Hyungjoon Kim\thanks{Corresponding author: hyungjoon@changwon.ac.kr}}

\affil[1]{Department of Computer Engineering,
Changwon National University,
Changwon-si, Gyeongsangnam-do 51140, Republic of Korea}

\date{}

\begin{document}

\maketitle

\begin{abstract}
Forecasting corporate tax avoidance proxies from firm--year panel data is challenging because predictive signals are distributed across short firm histories and related targets, while screening-oriented use requires transparent model behavior. We propose PaGNet (Panel-Aware GBDT--Neural Network), a two-branch hybrid that combines a LightGBM branch using panel-temporal summaries with a Panel-MLP branch using attention-pooled temporal aggregation and shared-trunk multi-task learning. A per-target validation-optimal blender produces both the final prediction and a compact branch-reliance diagnostic without trainable fusion parameters. On the KoTaP panel of 1{,}754 Korean listed firms from 2011--2024, PaGNet is evaluated under a leakage-free, shared-hyperparameter protocol across four feature regimes. In the direct-proxy-lag-excluded FS1 regime and the tax-history-augmented FS2 regime, accrual targets (TSTA, TSDA) route stably to the LightGBM branch, where PaGNet raises explained variance over the strongest of six baselines by roughly $0.08$--$0.11$ on the primary split. GETR often leans toward the neural branch, while CETR exposes a validation--test branch-selection mismatch rather than a stable branch assignment. A panel-flatten control shows that most accrual gains come from observed multi-year base-panel values, with PaGNet's panel-aware representation adding a smaller but directionally consistent refinement. Rolling-origin analysis confirms stable accrual routing, bounds ETR diagnostics to split-specific behavior, and identifies a far-horizon split where supervised models underperform naive persistence. PaGNet is therefore best viewed not as a universally superior tabular learner, but as a proxy-aware panel model that combines competitive forecasting with explicit per-target branch-reliance reporting.
\end{abstract}

\noindent\textbf{Keywords:}
Corporate tax avoidance, firm--year panel data,
GBDT--neural hybrid, leakage-free evaluation,
multi-task learning, panel-aware tabular learning.

\vspace{1em}

\section{Introduction}
\label{sec:introduction}

Since 2011, Korean listed firms have been required to adopt Korean International Financial Reporting Standards (K-IFRS), strengthening the formal consistency of financial reporting and inter-firm and inter-period comparability \cite{ref:fsc2007,ref:ifrs2013,ref:deloitte2013}. This regulatory transition has expanded the feasibility of constructing long-horizon firm--year panel datasets and applying quantitative prediction analyses to such data. Korea's Data Analysis, Retrieval and Transfer System (DART) has further established an XBRL-based disclosure infrastructure under K-IFRS, enhancing the accessibility and machine-readability of structured financial information \cite{ref:dart,ref:xbrlkorea}. Together, these developments have created an environment in which large-scale panel data can support both empirical analysis and data-driven monitoring or screening applications \cite{ref:shin2017,ref:hong2023,ref:jung2020}.

To address the need for a research-ready, standardized panel dataset, the recently released KoTaP (Korean Tax Avoidance Panel) dataset provides firm--year information for Korean non-financial listed companies covering 2011--2024, with 1{,}754 firms and 12{,}653 firm--year observations across 65 variables \cite{ref:kotap2026}. KoTaP centers on four tax avoidance proxy measures: two effective tax rate (ETR) variants, cash ETR (CETR) and GAAP ETR (GETR), and two book--tax difference (accrual-based) variants, the total book--tax difference scaled by total assets (TSTA) and its discretionary counterpart (TSDA). Forecasting these proxies one year ahead from currently observable information ($t \to t{+}1$) is of practical interest as a risk-screening tool. Doing so faithfully on a panel, however, is non-trivial: information leakage must be carefully prevented, and the predictive signal may be distributed across multi-year history and across related targets that conventional tabular learners are not designed to exploit.

\textbf{Accuracy is necessary but not sufficient: screening models must also be inspectable.} Tax authorities increasingly use data-driven models to prioritize returns for audit: the U.S. Internal Revenue Service already employs machine-learning models to select individual, refundable-credit, and large-partnership returns for examination \cite{ref:gao2024ai}, and the relevance of predicting tax outcomes such as the effective tax rate for investors, analysts, auditors, and regulators is well documented in the accounting literature \cite{ref:bogachek2026,ref:guenther2023}. Crucially, oversight bodies have emphasized that such models must remain transparent and explainable; the U.S. Government Accountability Office cautioned that without adequate documentation a tax authority ``risks not being able to explain how it selected taxpayers'' for audit \cite{ref:gao2024ai}. A screening model whose proxy-level rationale cannot be documented is difficult to act on, justify, or defend. This motivates models that expose their target-level branch reliance rather than emitting only an end-to-end black-box score. Our aim is therefore not to propose a universally superior tabular learner, but to take corporate tax avoidance as a concrete, economically grounded prediction problem and design a hybrid that is accurate where accuracy is achievable and structurally transparent about validation-selected proxy-level branch reliance.

Existing tabular learners fall into two families, neither designed for this setting. Gradient boosting decision trees, including XGBoost \cite{ref:xgboost}, LightGBM \cite{ref:lightgbm}, and CatBoost \cite{ref:catboost}, remain strong baselines but treat each sample as independent and identically distributed (IID) and each target independently. Recent deep tabular models such as TabNet \cite{ref:tabnet}, TabTransformer \cite{ref:tabtransformer}, FT-Transformer \cite{ref:ftt}, and ExcelFormer \cite{ref:excelformer} have advanced IID tabular benchmarks, but none handles the panel structure inherent in firm--year data or leverages the relatedness of multiple proxies; applied to a panel, they collapse the per-firm time structure to the most recent year and model each target alone. A separate line of GBDT--neural hybrids \cite{ref:lightgbm_nn,ref:gate,ref:tmlp,ref:gate_fusion,ref:sahgal_m5} combines the inductive biases of tree ensembles with neural representation learning and has shown that validation-optimized linear combinations of tree and neural predictors outperform single-family submissions in large-scale forecasting. These hybrids, however, are restricted to IID tabular data with a single target. Whether their principles transfer to panel data with multiple related targets, where each branch may exploit a different source of temporal or cross-target signal, and where the targets differ sharply in noise (high-autocorrelation accrual proxies versus noise-dominant ETR proxies), has received little systematic study.

We propose PaGNet (Panel-Aware GBDT--Neural Network), a hybrid two-branch model for multi-target tax avoidance forecasting on firm--year panel data. PaGNet combines a LightGBM branch operating on panel-temporal aggregate features over $K{=}3$ years of history with a Panel-MLP branch performing attention-pooled temporal aggregation and multi-task learning across the four targets. The two branches are fused through a per-target validation-optimal blender that adapts the branch allocation without introducing trainable fusion parameters. The central object of our study is not a single accuracy number but the per-target blend weight $\lambda^*_m$, which functions as a \emph{branch-reliance diagnostic}: it reports, for each proxy, whether validation performance allocates the output more strongly to tree-based panel-temporal aggregation or to the neural branch's shared representation. This readout is not firm-year-level feature attribution, but a compact target-level summary of branch reliance. On the primary evaluation split, this decomposition aligns most clearly with the accrual proxies: TSTA and TSDA route stably to the LightGBM branch, reflecting their strong temporal persistence. GETR often leans toward the Panel-MLP branch, while CETR exposes a validation--test branch-selection mismatch rather than a stable branch assignment. We are deliberately careful about how far this reading generalizes: a rolling-origin analysis over three temporal splits (Section~\ref{subsec:rolling}) shows that accrual routing is stable across origins, whereas ETR branch preferences are split-sensitive. We therefore treat the diagnostic as most reliable where the signal is strong, namely the accrual targets, and present the ETR reading as split-specific evidence rather than a general guarantee. In the direct-proxy-lag-excluded FS1 and FS2 regimes, PaGNet lifts accrual-target $R^2$ over the strongest of six baselines by $+0.08$ to $+0.11$ on the primary split.

To prevent these accuracy gains from being misattributed, we pair the model with a controlled panel-flatten experiment. The purpose of this control is to separate, as far as possible, two sources of gain that prior panel-tabular work often conflates: access to observed multi-year values and the representation used to process them. We grant all six IID baselines the same observed $K{=}3$-year base-panel values by concatenating the multi-year history along the feature axis, without changing their architectures. This does not reproduce the full PaGNet representation: the flat baselines do not receive PaGNet's explicit history-completeness metadata, engineered temporal summaries, or attention-based temporal pooling. The result is therefore best read as an observed-history control rather than a perfectly information-identical comparison. The finding is deliberately conservative for our own model: observed multi-year information explains most of the accrual-target lift, while PaGNet's panel-aware representation adds a smaller but directionally consistent refinement on top. We regard this attribution as a contribution rather than a weakness: it reduces over-attribution of gains to architecture alone and positions PaGNet's value where it is best supported, in proxy-level branch-reliance reporting and in its performance under direct-proxy-lag-excluded feature regimes.

The main contributions of this work are as follows.
\begin{enumerate}
    \item \textbf{A panel-aware, multi-target GBDT--neural hybrid whose blend weights provide a branch-reliance diagnostic.} We extend the IID GBDT--NN hybrid line \cite{ref:lightgbm_nn,ref:tmlp,ref:gate,ref:gate_fusion} to the panel multi-target setting through a per-target validation-optimal blender. The resulting weights expose branch--target heterogeneity not reported by the single-family baselines considered here: stable tree-branch routing for the accrual targets and split-sensitive behavior for the noisier ETR targets, a distinction we make explicit rather than gloss over.

    \item \textbf{An observed-history-versus-representation decomposition via panel-flatten control.} Through a controlled experiment in which six IID baselines receive the same observed $K{=}3$-year base-panel values, we estimate how much of the lift stems from multi-year information access versus PaGNet's panel-aware representation. We find that observed history carries most of the accrual-target lift, while branch design, temporal summaries, mask-aware processing, and attention-pooled aggregation add a smaller but directionally consistent refinement.

    \item \textbf{An evaluation methodology that locates where PaGNet's gains come from and where they fail.} A leakage-free protocol with shared hyperparameters reduces tuning-driven explanations; per-seed dispersion places each margin against its run-to-run variability; an empirical AR(1) ceiling analysis explains the FS3 and FS4 regimes in which strong autoregressive signal compresses architectural differences; and a rolling-origin analysis over three temporal splits separates stable findings (accrual routing and the ETR/accrual predictability bifurcation) from split-specific findings (the CETR branch-selection mismatch and the accrual lift, which holds on the near-origin splits but fails in the far-horizon stress split).
\end{enumerate}

The remainder of the paper is organized as follows. Section~\ref{sec:related} reviews related work in tabular deep learning, panel-data forecasting, GBDT--neural hybrid architectures, and tax avoidance prediction. Section~\ref{sec:method} formalizes the leakage-free forecasting problem, presents the PaGNet architecture, and describes the baseline models including the panel-flatten variants. Section~\ref{sec:experiments} reports results across the four feature sets, the panel-flatten analysis, the empirical AR(1) baseline analysis, and component ablations. Section~\ref{sec:conclusion} concludes with deployment implications and limitations.

\section{Related Work}
\label{sec:related}

\subsection{Tabular Deep Learning}
\label{subsec:tabular_dl}
Tabular data analysis remains dominated by tree-based ensembles, particularly gradient boosting decision trees (GBDTs), due to their robustness to heterogeneous feature scales, intrinsic handling of categorical variables, and strong out-of-the-box performance with limited tuning. XGBoost \cite{ref:xgboost} established the practical standard for scalable boosted trees through regularization, sparsity-aware optimization, and system-level engineering. LightGBM \cite{ref:lightgbm} extended this line via histogram-based splitting, gradient-based one-side sampling, and exclusive feature bundling, achieving high efficiency on large-scale high-dimensional data. CatBoost \cite{ref:catboost} introduced ordered target statistics for categorical variables together with permutation-driven debiasing, providing strong performance when categorical features dominate. These three methods constitute the canonical GBDT baselines used throughout the tabular learning literature.

Several deep tabular architectures have emerged to challenge GBDTs. TabNet \cite{ref:tabnet} uses sequential attention with sparse feature masks to combine performance with interpretability. TabTransformer \cite{ref:tabtransformer} applies multi-head self-attention to categorical embeddings to learn contextual representations, while continuous features are passed through a separate multi-layer perceptron path. FT-Transformer \cite{ref:ftt} extends this idea by tokenizing all features (both numerical and categorical) and applying a CLS-token-based transformer encoder, providing a strong general-purpose deep tabular baseline. Gorishniy \etal{} \cite{ref:ftt} systematically revisited these models and showed that, with careful tuning, deep tabular models can match GBDTs on many benchmarks. Most recently, ExcelFormer \cite{ref:excelformer} introduced semi-permeable attention (SPA), a one-way lower-triangular masking scheme over features ordered by training-set mutual-information importance, and interaction-aware initialization (IAI), an attenuated Kaiming scheme that stabilizes attention training; combined with gated tokenization and gated linear units, ExcelFormer reports state-of-the-art results on a 96-dataset tabular benchmark. None of these methods, however, is designed to handle the panel structure inherent in firm--year data or to leverage the relatedness of multiple economically related targets. In our experiments, we additionally evaluate TabTransformer, FT-Transformer, and ExcelFormer under a panel-flatten input that grants access to the same observed multi-year base-panel values, providing an observed-history-versus-representation comparison rather than a perfectly information-identical benchmark.

\subsection{Panel-Data and Time-Series Forecasting}
\label{subsec:panel_ts}
Classical time-series methods such as ARIMA and exponential smoothing offer well-understood baselines on individual series \cite{ref:hyndman_forecast,ref:hyndman_ets}, while Arellano--Bond dynamic panel generalized method of moments \cite{ref:arellano_bond} addresses endogeneity and temporal dependence in panels with short time horizons. With the rise of deep learning, several global forecasting architectures have been proposed to share parameters across many related series: DeepAR \cite{ref:deepar} formulates probabilistic forecasting with autoregressive recurrent networks; Deep State Space Models \cite{ref:dssm} combine state-space structure with deep parameterization; LSTNet \cite{ref:lstnet} integrates convolutional and recurrent components for multivariate series; and the Temporal Fusion Transformer (TFT) \cite{ref:tft} provides interpretable multi-horizon forecasting with static and time-varying covariates, introducing learned attention pooling over short histories that we adopt in the Panel-MLP branch of PaGNet. Chronopoulos \etal{} \cite{ref:chronopoulos} and Yang \etal{} \cite{ref:yang_panel} further analyze the use of deep networks in panel-data settings, focusing on fixed effects and cross-sectional dependence.

These methods, however, are primarily designed for forecasting where the time horizon is long relative to the cross-section. They are less directly applicable when the per-firm time series is short (in our case $K{=}3$ years of history per firm--year sample) and the predictive task is more tabular in nature, predicting next-year scalar targets from current and recent features. A further gap concerns evaluation: when multi-year history is supplied to a tabular or panel model, the resulting accuracy gain conflates two distinct effects, the additional information carried by the extra years and the model's architectural capacity to exploit temporal structure, and prior panel-tabular studies typically report only their sum without separating them. PaGNet occupies a different position in the design space: it treats the input as a short panel of tabular records, applies tree-based and attention-based aggregation in parallel rather than as a sequential autoregressive setup, fuses the two branches per target via validation-optimal blending, and is evaluated against an observed-history panel-flatten control (Section~\ref{subsec:baselines}) that helps separate access to multi-year base-panel values from PaGNet's panel-aware representation.

\subsection{Tax Avoidance Prediction and Risk Screening}
\label{subsec:tax_avoidance}
Tax avoidance is not directly observable and is conventionally measured by proxies derived from financial disclosures \cite{ref:hanlon_heitzman}. Effective tax rate (ETR) proxies, both cash-based (CETR) and GAAP-based (GETR), capture the ratio of tax burden to pre-tax income \cite{ref:dyreng2008,ref:hanlon_heitzman}, while book--tax differences (BTD) proxies capture discrepancies between accounting income and taxable income \cite{ref:desai_dharmapala,ref:hanlon_heitzman}. Frank \etal{} \cite{ref:frank2009} document a positive relationship between aggressive tax reporting and aggressive financial reporting.

A growing body of work applies machine learning to tax-related prediction. Guenther \etal{} \cite{ref:guenther2023} use machine learning to predict one-year-ahead ETR and compare predictive bias and precision across feature sets, showing that disclosure-based features carry substantial signal. Borrotti \etal{} \cite{ref:borrotti2023} predict aggressive tax positioning of European corporate groups from accounting information. Rahman \etal{} \cite{ref:rahman2020} apply logistic regression and tree-based models to detect tax avoidance in Malaysian listed firms. Related risk-screening literature has used financial-ratio-based scoring for earnings manipulation \cite{ref:beneish}, accounting misstatement risk \cite{ref:dechow2011}, bankruptcy \cite{ref:altman}, and text-based event prediction \cite{ref:cecchini}.

A parallel concern in this application domain is that screening models must be not only accurate but also transparent and auditable. In credit scoring, regulators demand transparent and auditable models, which has kept simple learners such as logistic regression and decision trees in widespread use and has motivated frameworks for rendering ``black box'' models transparent, auditable, and explainable while retaining their predictive power \cite{ref:bucker2022}. The same tension appears directly in tax administration: studies of machine-learning-based audit selection at the U.S. Internal Revenue Service emphasize that algorithmic allocation decisions carry equity and accountability stakes and must be examinable rather than opaque \cite{ref:black2022}, echoing the oversight guidance that an authority must be able to explain how it selected taxpayers for audit \cite{ref:gao2024ai}. These works establish the practical relevance of machine learning for tax-related prediction but typically use IID classifiers or regressors on aggregated firm--year observations, without explicitly modeling the panel structure or the relatedness of multiple proxies, and without reporting target-level model-family reliance. Building on the KoTaP dataset and its leakage-free $t \to t{+}1$ protocol \cite{ref:kotap2026}, our work contributes a panel-aware hybrid architecture that explicitly leverages panel structure and multi-target relatedness, exposes a per-target branch-reliance readout, conducts a unified benchmark spanning gradient boosting, IID deep tabular, panel-flatten deep tabular, and hybrid families under a shared default-hyperparameter protocol, and provides component-level ablations that justify each architectural choice.

\subsection{GBDT--Neural Hybrid Architectures}
\label{subsec:hybrid}

A growing line of recent work has explored hybrid architectures that combine the inductive biases of gradient boosting with the representation learning of neural networks, motivated by the observation that GBDTs and DNNs excel on different subsets of tabular tasks and that their errors are often partially uncorrelated. Existing hybrids fall into three main lines: output-level blending, joint-training architectures with tree-style inductive biases, and tree-guided neural representation learning.

Output-level blending combines independently trained GBDT and neural predictors at the output. Zhang \cite{ref:lightgbm_nn} concatenates LightGBM and MLP predictions for high-frequency realized volatility forecasting. In the M5 forecasting competition, top-ranked submissions blend gradient-boosted trees and neural networks through validation-optimized linear weights, finding that no single family dominates and that adaptive per-series blending captures complementary signals \cite{ref:sahgal_m5}. PaGNet's validation-optimal blender follows this line with two extensions: per-target rather than per-series blending, motivated by noise heterogeneity across CETR, GETR, TSTA, and TSDA, and an additional shrinkage parameter toward the training mean that stabilizes RMSE on volatile targets.

A second line embeds tree-like structures directly into a differentiable neural backbone. GATE \cite{ref:gate} integrates differentiable gated decision trees with a GRU-style feature gating mechanism, while NODE \cite{ref:node} proposes oblivious decision ensembles as a differentiable building block. These approaches retain a single end-to-end trainable model rather than maintaining two separate branches.

A third line uses GBDTs to inform the architecture or training of a neural network. T-MLP \cite{ref:tmlp} uses GBDT feature importance as a sparse gating mask on a randomly initialized MLP and ensembles three branches with different learning rates. GATE-Fusion \cite{ref:gate_fusion} extracts leaf-index embeddings from multiple GBDTs and feeds them with the original features into an MLP fusion head for credit risk prediction. Unlike validation-optimal blending, these approaches require joint training of the tree and neural components and introduce fusion parameters that must be regularized.

The three lines above all operate on IID tabular data with a single target, and they treat the fusion mechanism primarily as an accuracy device: the learned combination weights are not read back as a target-level diagnostic of model-family reliance. To our knowledge, no prior work has extended hybrid GBDT--neural architectures to the panel multi-target setting where each sample is a short multi-year history rather than a single row, multiple economically related targets are predicted jointly, and the fusion weights themselves serve as a diagnostic. PaGNet contributes four design choices specific to this setting: an explicit panel-temporal aggregate representation for the LightGBM branch summarizing the $K{=}3$-year history through statistical moments and lag deltas; a Panel-MLP branch with attention-pooled temporal aggregation, gated last-step fusion, and a shared multi-task representation across the four targets; a per-target validation-optimal blender that adapts the branch allocation separately for each target; and the use of that blender's per-target weights as a branch-reliance diagnostic that reports whether validation performance allocates each target output more strongly to tree-based temporal aggregation or to cross-target neural representation sharing.

\section{Methodology}
\label{sec:method}

\begin{figure*}[t!]
    \centering
    \includegraphics[width=\textwidth]{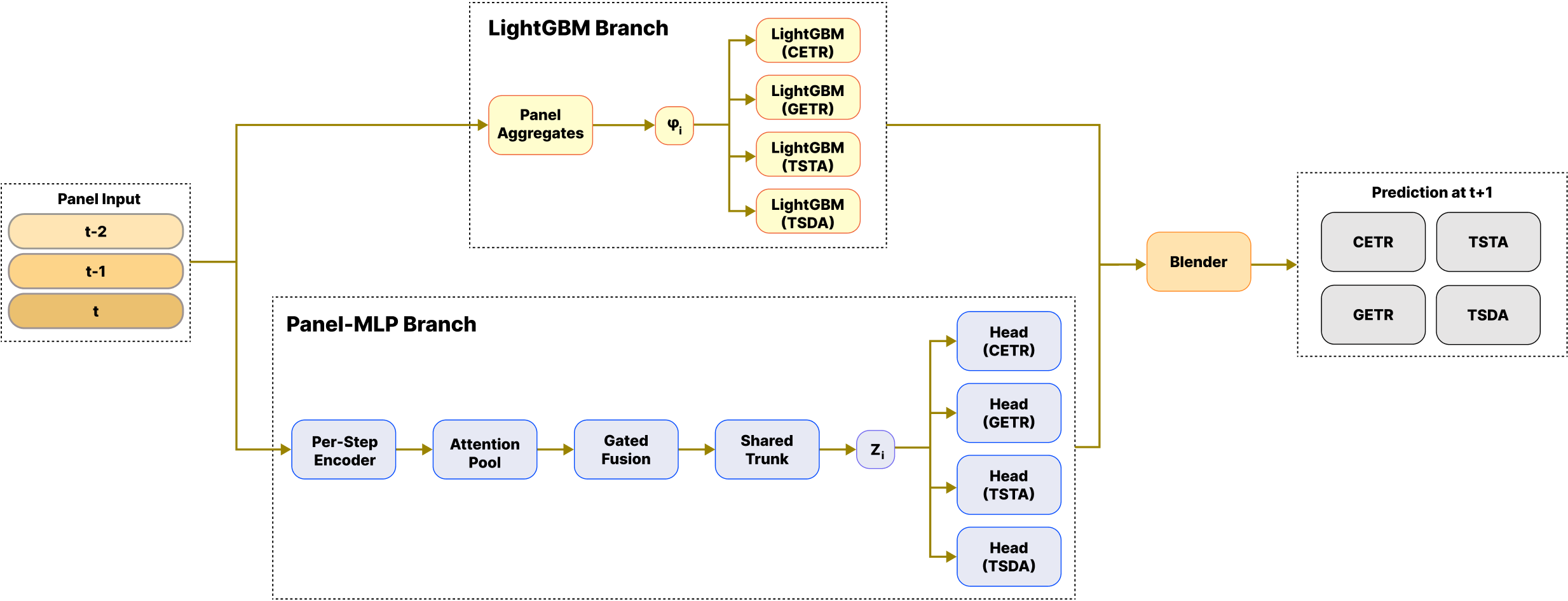}
    \caption{PaGNet architecture overview. Two parallel branches (LightGBM and Panel-MLP) process the $K{=}3$-year firm panel, and a per-target validation-optimal blender fuses their predictions for the four targets. Component details are described in Section~\ref{subsec:pagnet}.}
    \label{fig:architecture}
\end{figure*}

This section formalizes the leakage-free one-year-ahead forecasting task, describes the KoTaP panel dataset, presents our proposed PaGNet architecture in detail, and summarizes the baseline models used for comparison, including the panel-flatten variants introduced to separate observed multi-year history from PaGNet's panel-aware representation. We follow the dataset construction and evaluation protocol established in the original KoTaP paper \cite{ref:kotap2026}, which we summarize here for completeness, and then focus on the novel architectural contributions of PaGNet.

\subsection{Problem Formulation and KoTaP Dataset}
\label{subsec:problem}

We formulate the one-year-ahead prediction task on a firm--year panel as follows. Let $i \in \{1, \ldots, N\}$ index firms and $t$ index fiscal years. For each firm--year observation, KoTaP provides a feature vector $X_{i,t} \in \mathbb{R}^d$ and a target vector $Y_{i,t} = [\text{CETR}, \text{GETR}, \text{TSTA}, \text{TSDA}]_{i,t}^\top \in \mathbb{R}^4$. The task is to predict $Y_{i,t+1}$ from information available no later than year $t$.

KoTaP \cite{ref:kotap2026} contains 12{,}653 firm--year observations from 1{,}754 Korean listed non-financial firms over 2011--2024. The panel is unbalanced: firms enter and exit the sample at different times due to listings, delistings, and reporting gaps. Variables are organized into identifier and meta fields (\eg{} \texttt{stock}, \texttt{year}, \texttt{ind}), raw accounting items extracted directly from disclosures (\eg{} \texttt{asset}, \texttt{sales}, \texttt{tax}, \texttt{ocf}), derived financial ratios and lag features (\eg{} \texttt{ROA}, \texttt{LEV}, \texttt{SIZE}, \texttt{lag\_asset}), four tax avoidance proxy targets (CETR, GETR, TSTA, TSDA), and auxiliary tax proxies (CETR3, GETR5, A\_CETR, etc.). Full variable definitions follow KoTaP \cite{ref:kotap2026}.

To support panel modeling, we construct supervised samples as $(\mathcal{H}_{i,t}, Y_{i,t+1})$ where $\mathcal{H}_{i,t} = \{X_{i,t-K+1}, \ldots, X_{i,t}\}$ is the firm's $K$-year history ending at year $t$. We use $K{=}3$ throughout this work, which balances temporal context with the need to retain firms that have short observation windows: under the unbalanced KoTaP panel, larger $K$ would discard a growing fraction of firm--year samples that lack a full history, while $K{=}1$ recovers the IID year-$t$ setting that the baselines already represent. We fix $K{=}3$ for all models rather than tuning it per model, and treat sensitivity to the window length $K$ as a question for future work (Section~\ref{sec:conclusion}); the panel-flatten control in Section~\ref{subsec:baselines} grants every baseline the same $K{=}3$ history, so the window length is held constant across the architecture comparison and does not confound it. When fewer than $K$ years are available (\eg{} for firms with recent listings or earlier gaps), the missing year slots are zero-padded and a binary mask $m \in \{0,1\}^K$ indicates valid years; the mask is supplied to the temporal attention layer so that padded years do not contribute to attention scores. In the training split, $61.3\%$ of pairs have full $K{=}3$ coverage, with an average of $2.42$ observed years per window. Pairs $(\mathcal{H}_{i,t}, Y_{i,t+1})$ are constructed only when both year $t$ and year $t{+}1$ are observed for firm $i$. The final sample counts are 6{,}502 (train), 879 (validation), and 929 (test).

To quantify the contribution of different variable groups, we evaluate four feature configurations following the KoTaP protocol. \textbf{FS1} (target-proxy-excluded baseline) uses raw and/or derived variables but excludes all tax avoidance proxies and their aggregates; we report three sub-configurations: Raw-only, Derived-only, and Raw+Derived. \textbf{FS2} adds KoTaP's ten auxiliary tax aggregates, taken as released, which summarize multi-year and benchmark-adjusted ETR behavior using only information available up to year $t$: the three- and five-year sum-over-sum cumulative cash and GAAP ETRs (CETR3, CETR5, GETR3, GETR5) and their industry--size-adjusted counterparts (A\_CETR, A\_GETR, A\_CETR3, A\_GETR3, A\_CETR5, A\_GETR5), all listed in Table~\ref{tab:app_vardict}. Following KoTaP \cite{ref:kotap2026}, the adjusted variables subtract the mean ETR of firms in the same industry and asset-size quartile in year $t$, and the long-run ETRs use the sum-over-sum convention over rolling windows ending at year $t$. \textbf{FS3} adds the lagged target proxies $Y_{i,t}$ (\ie{} $\text{CETR}_{i,t}$, $\text{GETR}_{i,t}$, $\text{TSTA}_{i,t}$, $\text{TSDA}_{i,t}$) as input features. \textbf{FS4} adds both auxiliary aggregates and lagged target proxies. All variants use Raw+Derived as the base configuration to remain consistent with KoTaP \cite{ref:kotap2026}. By construction, every feature in every configuration is computed from information dated no later than year $t$, so the target $Y_{i,t+1}$ never enters the input directly or indirectly through any aggregate.

\subsection{Leakage-Free Evaluation Protocol}
\label{subsec:protocol}

We adopt the leakage-free evaluation protocol of KoTaP \cite{ref:kotap2026}, which enforces three principles throughout: ex-ante information availability, temporal ordering, and separation of fit and evaluation.

Inputs are restricted to variables observable at year $t$. The target $Y_{i,t+1}$ is never used as input. Derived variables (ratios, lag features) are computed using only values up to year $t$. The FS2 auxiliary aggregates are the KoTaP-released long-run and industry--size-adjusted ETRs, each computed from information dated no later than year $t$: rolling windows ending at $t$ for the cumulative ETRs, and the year-$t$ industry--size cross-section for the adjusted variables, with no observation from year $t+1$ or later entering any aggregate. Because every such quantity is dated at or before year $t$, it is observable at prediction time and introduces no leakage of the year-$t{+}1$ target. The FS3 and FS4 lagged target proxies use only $Y_{i,t}$, never $Y_{i,t+1}$.

We split the data by input year $t$ as our primary evaluation protocol: training uses $t \in [2011, 2019]$, validation uses $t{=}2021$ (target 2022), and test uses $t{=}2023$ (target 2024). Buffer years ($t{=}2020$ and $t{=}2022$) are excluded as supervised prediction origins (no training, validation, or test sample uses 2020 or 2022 as its input year $t$), which reduces temporal overlap and gives a more conservative out-of-time evaluation; they remain available as historical covariates inside a firm's $K{=}3$ window when observable at the prediction origin (for example, 2022 can appear in the $t{=}2023$ test history). This single fixed split follows the KoTaP reference protocol to keep our results directly comparable; we comment on robustness to alternative temporal splits in Section~\ref{sec:experiments}.

All preprocessing operations are fit on the training split only and applied (\texttt{transform}) to validation and test. Numerical features are first winsorized to the train-set $0.5$th and $99.5$th percentiles and then standardized using train-set means and standard deviations. Categorical features (industry code, 47 classes) are integer-encoded, with an out-of-vocabulary index for codes unseen in training. Missing years within a firm's $K{=}3$ history are handled by zero-padding together with a binary validity mask supplied to both branches, rather than by mean imputation, so padded slots do not contribute to temporal aggregates or attention scores. The FS3/FS4 lagged target proxies are the only inputs with residual missing values; these are imputed with the train-set target mean, standardized, and clipped to $\pm 3$ standard deviations. For the Panel-MLP branch the four targets are standardized with train-set target moments and predictions are de-standardized before scoring, and the Huber loss ($\delta=0.05$) is therefore evaluated on the standardized target scale; the LightGBM branch is fit directly on the raw target scale. Hyperparameter selection and early stopping use the validation split; final test performance is reported once after model selection.

\subsection{PaGNet Architecture}
\label{subsec:pagnet}

PaGNet consists of two parallel branches, a LightGBM branch on panel-temporal aggregate features and a Panel-MLP branch with attention-pooled temporal aggregation and multi-task heads, fused through a per-target validation-optimal blender (Fig.~\ref{fig:architecture}). The remainder of this section formalizes each component.

\subsubsection{Overview and Notation}
\label{subsubsec:overview}

The input to PaGNet is the $K{=}3$-year panel introduced in Section~\ref{subsec:problem}: for each supervised sample $(\mathcal{H}_{i,t}, Y_{i,t+1})$, we have a numerical tensor $X^{\text{num}}_{i} \in \mathbb{R}^{K \times F_n}$, a binary tensor $X^{\text{bin}}_i \in \{0,1\}^{K \times F_b}$, a categorical industry vector $c_i \in \{1, \ldots, 47\}^{K}$, a validity mask $m_i \in \{0,1\}^{K}$, and a 4-target vector $Y_i \in \mathbb{R}^4$ for $\{\text{CETR}, \text{GETR}, \text{TSTA}, \text{TSDA}\}$ at $t{+}1$. We denote by $k_i^{\text{last}} \in \{0, \ldots, K{-}1\}$ the index of the last valid timestep, which equals $K{-}1$ for full-history firms and a smaller index for short-history firms with padded leading slots.

\subsubsection{LightGBM Branch with Panel-Temporal Feature Engineering}
\label{subsubsec:lgbm_branch}

The LightGBM branch consumes a flat feature vector $\phi_i \in \mathbb{R}^{D_\phi}$ that summarizes the $K$-year panel history. For each numerical feature $j$, we compute seven per-sample aggregates over the valid timesteps $\mathcal{V}_i = \{k : m_{i,k} = 1\}$: mean $\mu_{i,j}$, standard deviation $\text{std}_{i,j}$, minimum, maximum, range, last-vs-prior-step delta $\Delta^{\text{prev}}_{i,j} = X^{\text{num}}_{i, k_i^{\text{last}}, j} - X^{\text{num}}_{i, k_i^{\text{prev}}, j}$, and last-vs-mean delta $\Delta^{\text{mean}}_{i,j} = X^{\text{num}}_{i, k_i^{\text{last}}, j} - \mu_{i,j}$, where $k_i^{\text{prev}}$ is the immediately prior valid timestep ($\text{std}_{i,j}$ is set to zero when $|\mathcal{V}_i| < 2$). The full feature vector concatenates the last-step raw values, the seven aggregates, and mask metadata:
\begin{equation}
\begin{aligned}
    \phi_i = \big[\,&X^{\text{num}}_{i, k_i^{\text{last}}, :},\; X^{\text{bin}}_{i, k_i^{\text{last}}, :},\; c_{i, k_i^{\text{last}}}, \\
                    &\mu_{i,:},\; \text{std}_{i,:},\; \text{max}_{i,:},\; \text{min}_{i,:},\; \text{range}_{i,:}, \\
                    &\Delta^{\text{prev}}_{i,:},\; \Delta^{\text{mean}}_{i,:},\; n^{\text{valid}}_i,\; \mathbb{1}[|\mathcal{V}_i| \geq 2] \,\big].
\end{aligned}
\label{eq:phi_concat}
\end{equation}
This gives LightGBM comparable multi-year information to the Panel-MLP branch but in a form natively consumable by tree splits. For FS3 and FS4, the standardized lagged target proxy $\tilde{Y}_{i,t}$ is appended as four additional columns.
 
For each target $m \in \{\text{CETR}, \text{GETR}, \text{TSTA}, \text{TSDA}\}$, we train an independent LightGBM regressor $f^{\text{lgb}}_m: \mathbb{R}^{D_\phi} \to \mathbb{R}$ with squared error loss and default hyperparameters shared with all GBDT baselines (3{,}000 boosting rounds with early stopping on validation RMSE, learning rate $0.05$, 63 leaves, minimum 20 samples per leaf, feature and bagging fraction $0.9$, $\ell_2$ regularization $1.0$). The industry code is treated as a native categorical feature.

\subsubsection{Panel-MLP Branch}
\label{subsubsec:nn_branch}

The Panel-MLP branch operates on the panel tensor $(X^{\text{num}}_i, X^{\text{bin}}_i, c_i, m_i)$ and jointly produces a 4-dimensional prediction through a shared trunk and four target-specific heads. The branch encodes each timestep, aggregates across time, and feeds a shared trunk that drives target-specific heads. Feature-set additions enter this panel tensor rather than a separate covariate path: the FS2 auxiliary aggregates are time-varying and occupy their own channels of $X^{\text{num}}_i$ at every timestep (each year $k$ carries its own rolling value), whereas the FS3/FS4 lagged target proxies $\tilde{Y}_{i,t}$, being constant within a sample, are broadcast across all $K$ timesteps and concatenated to $X^{\text{num}}_i$ as four additional channels. The branch has no dedicated static-covariate encoder, so a sample-constant feature such as the lagged proxy is simply re-encoded identically at each step and then pooled with the rest.

The encoder first embeds the categorical industry code through a learnable lookup ($d_{\text{emb}}=8$). For each timestep $k$, numerical, binary, and categorical components are concatenated and passed through a residual-friendly MLP with layer normalization and GELU activation (hidden dimension $d=64$, dropout 0.1). A learnable year positional bias is added and padded timesteps are zeroed out via the validity mask. To produce a single $d$-dimensional summary of the $K$-year history, we apply multi-head attention ($H_{\text{attn}}=4$) with a learnable query vector $q \in \mathbb{R}^d$ shared across samples:
\begin{equation}
    p_i = \text{MHA}\big(q,\; \tilde{h}_i,\; \tilde{h}_i;\; \text{mask}\big) \in \mathbb{R}^d.
    \label{eq:attn_pool}
\end{equation}
The learned query allows attention weights based on content rather than position alone.

Since the most recent year is particularly informative for short-horizon forecasting, we fuse the attention-pooled summary $p_i$ with the last-valid-step encoding $\tilde{h}_{i, k_i^{\text{last}}}$ through a learned softmax gate. Let $[g_{i,0},\, g_{i,1}] = \mathrm{softmax}\big(W_g\,[\,p_i\,;\,\tilde{h}_{i, k_i^{\text{last}}}\,] + b_g\big)$ with learnable $W_g \in \mathbb{R}^{2 \times 2d}$ and $b_g \in \mathbb{R}^2$, so that $g_{i,0} + g_{i,1} = 1$; the fused representation is
\begin{equation}
    z^{\text{fuse}}_i = g_{i,0} \cdot p_i + g_{i,1} \cdot \tilde{h}_{i, k_i^{\text{last}}}.
    \label{eq:fused}
\end{equation}
The fused representation passes through a residual MLP trunk yielding $z_i \in \mathbb{R}^d$, from which four parallel target-specific heads $H_m: \mathbb{R}^d \to \mathbb{R}$ produce the joint prediction $\hat{Y}^{\text{nn}}_{i,m} = H_m(z_i)$ for $m \in \{\text{CETR, GETR, TSTA, TSDA}\}$. Predictions are de-standardized using training-set target moments.

The four targets are trained jointly through the shared trunk and target-specific heads, minimizing the equal-weight average of the four per-target Huber losses \cite{ref:huber} ($\delta=0.05$). We deliberately keep this multi-task objective parameter-free (the four losses are averaged with fixed unit weights rather than reweighted by learnable per-target scales), so that no target-weighting hyperparameters are introduced or tuned; the cross-target coupling that benefits the noisy ETR proxies comes entirely from the shared representation, not from a weighting scheme. We train for up to 100 epochs with AdamW (learning rate $3 \times 10^{-4}$, weight decay $10^{-5}$), batch size 128, OneCycle learning-rate schedule, and early stopping on validation 4-target average $R^2$ with patience 15. Test predictions are averaged across 5 random seeds.

\subsubsection{Validation-Optimal Blender}
\label{subsubsec:blend}

Let $\hat{Y}^{\text{lgb}}_{i,m}$ and $\hat{Y}^{\text{nn}}_{i,m}$ denote the LightGBM-branch and Panel-MLP-branch predictions for target $m$. The blended prediction is
\begin{equation}
    \hat{Y}^{\text{blend}}_{i,m} = (1 - s_m) \big[ \lambda_m \,\hat{Y}^{\text{lgb}}_{i,m} + (1 - \lambda_m) \,\hat{Y}^{\text{nn}}_{i,m} \big] + s_m \,\bar{Y}^{\text{train}}_m,
    \label{eq:blend}
\end{equation}
where $\lambda_m \in [0, 1]$ is the branch-reliance weight (the share placed on the tree branch), $s_m \in [0, 0.4]$ is a shrinkage coefficient toward the training-set target mean $\bar{Y}^{\text{train}}_m$, and both are selected per target via 2D grid search over $\lambda \in \{0, 0.05, \ldots, 1\}$ and $s \in \{0, 0.05, \ldots, 0.4\}$ to minimize the validation MSE. The grid search introduces no parameters trained on the test set.

The blender simultaneously (i) adapts the relative reliance on each branch per target, allowing heterogeneous noise structures across CETR, GETR, TSTA, and TSDA to receive different fusion weights; (ii) pulls predictions toward the training mean through $s_m$ on targets for which validation indicates that such regularization reduces error; and (iii) is differentiable-free, requiring no joint optimization or fusion-network parameters \cite{ref:gate_fusion}.

Beyond its role in fusion, the per-target weight $\lambda^*_m$ doubles as a \emph{branch-reliance diagnostic}. Because $\lambda^*_m$ is a single validation-optimal share placed on the tree branch for target $m$, held constant across all firm--year samples rather than varying per observation, its value reads out, for each proxy as a whole, whether that target's predictions are on aggregate better explained by tree-based panel-temporal aggregation ($\lambda^*_m \to 1$) or by the cross-target representation sharing of the multi-task Panel-MLP ($\lambda^*_m \to 0$). A single trained pipeline thus emits, alongside its four predictions, a compact 4-row $(\lambda^*_m, s^*_m)$ table that reports, per target, which branch the proxy relies on; this is a target-level summary of branch reliance, not a firm-year-level or per-sample feature attribution, and it is a readout the single-family baselines we study do not produce. Section~\ref{subsec:ablation} reports the empirical $\lambda^*_m$ distribution and shows that it aligns with the economic and statistical structure of the four proxies. PaGNet's inference path thus decomposes cleanly into ``train each branch separately, fit the blender on validation, predict on test.''

\subsubsection{End-to-End Summary}
\label{subsubsec:e2e_summary}

The full PaGNet prediction substitutes the trained branch functions $f^{\text{lgb}}_m(\phi_i)$ and $H_m(z_i)$ into the blender (Eq.~\ref{eq:blend}) with validation-optimal $(\lambda_m^*, s_m^*)$. Unlike pure transformer panel models requiring a single end-to-end backbone, PaGNet decomposes into three independently inspectable artifacts: a LightGBM forest per target (with native feature-importance readouts on the panel-temporal aggregates), a single Panel-MLP backbone with attention-pool weights over the $K$-year history, and a 4-row $(\lambda_m^*, s_m^*)$ blend table that serves as the branch-reliance diagnostic of Section~\ref{subsubsec:blend}. This decomposability is what lets each of the four target predictions be attributed to an identifiable branch, and it underlies the deployment guidance in Section~\ref{subsec:discussion}.

\subsection{Baseline Models}
\label{subsec:baselines}

We compare PaGNet against six baselines spanning gradient boosting and deep tabular learning, plus six panel-flatten variants of those baselines that grant equal access to $K{=}3$ years of history. All baselines are trained under the same leakage-free protocol described in Section~\ref{subsec:protocol}: identical train/validation/test splits, identical preprocessing (\texttt{fit} on train only), identical 5-seed averaging, and, critically for fair comparison, identical default hyperparameter budgets per model family rather than per-target tuning. Categorical features (industry) are integer-encoded; LightGBM and CatBoost handle categoricals natively, while XGBoost and the deep tabular models treat them via embeddings or numeric encoding. We deliberately follow the standard practice of each model family rather than imposing a single uniform encoding, so that every baseline is evaluated in the form in which it is normally deployed; forcing a non-native encoding on any family would handicap it for reasons unrelated to the panel question we study.
 
The gradient boosting baselines are XGBoost \cite{ref:xgboost}, LightGBM \cite{ref:lightgbm}, and CatBoost \cite{ref:catboost}, the canonical GBDT methods for tabular data. We use a single default configuration shared across all four targets (learning rate $0.05$, depth $6$, 63 leaves where applicable, $\ell_2$ regularization $1.0$, feature and bagging fraction $0.9$, early stopping with 50-round patience on validation RMSE) and report mean test performance across 5 seeds. Tree-based baselines use only the year-$t$ slice of the panel as input (panel structure collapsed to IID); each target requires a separately trained model.
 
The deep tabular baselines are TabTransformer \cite{ref:tabtransformer}, FT-Transformer \cite{ref:ftt}, and ExcelFormer \cite{ref:excelformer}. TabTransformer applies multi-head self-attention to categorical tokens while passing numerical features through an MLP path. FT-Transformer tokenizes all features and processes them with a CLS-token-based transformer encoder. ExcelFormer adds gated tokenization, semi-permeable attention (SPA), interaction-aware initialization (IAI), and a GLU feed-forward block. All three are configured at embedding dimension $d{=}64$, $L{=}3$ layers, $H{=}8$ attention heads, dropout $0.1$, and an AdamW optimizer with learning rate $3 \times 10^{-4}$ and weight decay $10^{-5}$, matching the Panel-MLP branch of PaGNet for fair comparison, and trained on the year-$t$ slice ($K{=}1$). Each target is trained as a separate single-task model.
 
We also report two PaGNet ablation variants that hold architecture and protocol fixed. PaGNet-LGBM uses only the LightGBM branch ($f^{\text{lgb}}_m(\phi_i)$ directly, no blending), and PaGNet-NN uses only the Panel-MLP branch ($H_m(z_i)$ directly, no blending). The comparison among the full PaGNet and these two ablations quantifies each branch's isolated contribution, the blender's marginal gain, and the per-target reliance profile $\lambda_m^*$ (Section~\ref{subsec:ablation}). For these two branch-only variants we report RMSE, MAE, and $R^2$ computed from the corresponding pre-blend branch predictions, so each branch is evaluated on the same footing as the blended PaGNet.
 
To separate access to observed multi-year values from PaGNet's panel-aware representation, we additionally evaluate all six baselines under a panel-flatten configuration that concatenates the $K{=}3$ years of features along the feature axis. The industry code (essentially time-invariant) is included only from year $t$; FS2 auxiliary tax aggregates and FS3/FS4 lagged target proxies are appended as single year-$t$ columns. No architectural changes are made; only the input dimension grows by a factor of $K$. Missing history years are zero-padded, and the panel-flatten baselines do not receive the explicit validity mask or valid-year count that PaGNet's two branches use; this history-completeness metadata is therefore part of PaGNet's panel-aware design rather than shared information, and any small edge it confers is attributed to architecture rather than to additional input years. This gives the IID baselines closely comparable multi-year feature values to PaGNet without any architectural mechanism for explicit temporal modeling, and thereby serves as an \emph{observed-history} control rather than a perfectly information-identical comparison: any performance difference that remains between a panel-flatten baseline and PaGNet is attributable primarily to PaGNet's panel-aware representation rather than to access to additional years, up to the small input asymmetries noted in Section~\ref{subsec:panel_flatten}, which is the comparison that helps separate our representational contribution from the confound of observed-history access. The panel-flatten LightGBM differs from PaGNet-LGBM in that it concatenates raw lag columns rather than engineered temporal aggregates, allowing us to compare which form of multi-year representation tree-based learners benefit from. We restrict panel-flatten experiments to FS1 and FS2 because, as Section~\ref{subsec:fs4} shows, FS3/FS4 performance is dominated by lagged-target autocorrelation regardless of panel access.
 
The six standard models span three regimes (pure GBDT IID, IID deep tabular, panel-aware hybrid), the six panel-flatten variants provide an observed-history baseline, and the three PaGNet variants decompose the hybrid into its constituents (Table~\ref{tab:model_summary}).

\begin{table}[!t]
\centering
\caption{Summary of model variants. $K$: number of input years; Branches: GBDT, neural, or both; MT: multi-task joint training of the four targets; Blend: per-target validation-optimal fusion.}
\label{tab:model_summary}
\setlength{\tabcolsep}{3pt}
\renewcommand{\arraystretch}{1.1}
\footnotesize
\begin{tabular}{@{}lcccc@{}}
\toprule
Model & $K$ & Branches & MT & Blend \\
\midrule
XGBoost / LightGBM / CatBoost (IID)       & 1 & GBDT          & --        & --        \\
TabT / FT-Trans / ExcelFormer (IID)       & 1 & Neural        & --        & --        \\
XGBoost / LightGBM / CatBoost (pflat)     & 3 & GBDT          & --        & --        \\
TabT / FT-Trans / ExcelFormer (pflat)     & 3 & Neural        & --        & --        \\
\midrule
PaGNet-LGBM (ours)                        & 3 & GBDT          & --        & --        \\
PaGNet-NN (ours)                          & 3 & Neural        & \checkmark & --       \\
PaGNet (full, ours)                       & 3 & GBDT+Neural   & \checkmark & \checkmark \\
\bottomrule
\end{tabular}
\end{table}

\section{Experiments}
\label{sec:experiments}

This section reports our empirical evaluation on KoTaP. Section~\ref{subsec:setup} describes the experimental protocol with default-hyperparameter constraints shared across all models. Sections~\ref{subsec:fs1} and \ref{subsec:fs2} present the main results across FS1 (target-proxy-excluded, with three input sub-configurations) and FS2 (auxiliary tax aggregates), with brief data-side interpretations woven into each. Section~\ref{subsec:panel_flatten} reports the panel-flatten analysis that separates observed multi-year history from PaGNet's panel-aware representation, with full RMSE, MAE, and $R^2$ metrics. Section~\ref{subsec:fs4} presents FS3 and FS4 results with an empirical autoregressive ceiling analysis.

\subsection{Experimental Setup}
\label{subsec:setup}

All six baselines and three PaGNet variants share an identical leakage-free protocol (Section~\ref{subsec:protocol}) and, critically for fair comparison, identical default hyperparameter configurations per model family. Gradient boosting baselines (XGBoost, LightGBM, CatBoost) and the LightGBM branch of PaGNet use a single shared configuration across all four targets: 3{,}000 maximum boosting rounds with 50-round early stopping on validation RMSE, learning rate $0.05$, 63 leaves where applicable, minimum 20 samples per leaf, feature and bagging fractions both $0.9$, and $\ell_2$ regularization $1.0$. Deep tabular baselines (TabTransformer, FT-Transformer, ExcelFormer) and the Panel-MLP branch of PaGNet use embedding dimension $d{=}64$, $L{=}3$ layers, $H{=}8$ attention heads, dropout $0.1$, AdamW optimizer with learning rate $3 \times 10^{-4}$, weight decay $10^{-5}$, batch size 128, gradient clipping at norm $1.0$, OneCycle learning-rate schedule with 10\% warmup and cosine anneal, up to 100 epochs with 15-epoch early-stopping patience on validation RMSE (the multi-task variant uses the averaged 4-target $R^2$ instead). The PaGNet validation-optimal blender uses a $21 \times 9$ grid search over $\lambda \in [0, 1]$ and $s \in [0, 0.4]$ at intervals of $0.05$. Industry is treated as a categorical feature for LightGBM and CatBoost (native support), and as numerical or embedded for XGBoost and the deep tabular models. For each of the 4 feature sets, 4 targets, and 5 seeds in $\{42, 43, 44, 45, 46\}$, we report mean RMSE, MAE, and $R^2$ on the held-out test split (year 2024 targets).

\subsection{FS1: Target-Proxy-Excluded Baseline}
\label{subsec:fs1}

\begin{table*}[!ht]
\centering
\caption{FS1 (target-proxy-excluded) test performance across three input sub-configurations (mean over 5 seeds). Best per (target, sub-config, metric) in \textbf{bold}, second \underline{underlined}. Ranking decided at 4-decimal precision; ties at 4 decimals are broken at the 5th decimal so exactly one bold and one underline appear per cell. PaGNet-LGBM/-NN are the LightGBM/Panel-MLP branches in isolation.}
\label{tab:fs1}
\setlength{\tabcolsep}{3pt}
\footnotesize
\begin{tabular}{llccc|ccc|ccc}
\toprule
\multirow{2}{*}{Target} & \multirow{2}{*}{Model} & \multicolumn{3}{c|}{Raw-only} & \multicolumn{3}{c|}{Derived-only} & \multicolumn{3}{c}{Raw+Derived} \\
& & RMSE & MAE & $R^2$ & RMSE & MAE & $R^2$ & RMSE & MAE & $R^2$ \\
\midrule
\multirow{9}{*}{CETR}
& XGBoost & 0.2347 & 0.1862 & $-$0.1250 & 0.2341 & 0.1860 & $-$0.1184 & 0.2313 & 0.1836 & $-$0.0921 \\
& LightGBM & 0.2337 & 0.1863 & $-$0.1152 & 0.2329 & 0.1845 & $-$0.1069 & 0.2325 & 0.1850 & $-$0.1031 \\
& CatBoost & \underline{0.2288} & 0.1817 & \underline{$-$0.0688} & 0.2292 & 0.1813 & $-$0.0728 & 0.2278 & 0.1808 & $-$0.0592 \\
& TabTransformer & 0.2338 & 0.1863 & $-$0.1155 & 0.2327 & 0.1849 & $-$0.1055 & 0.2342 & 0.1866 & $-$0.1197 \\
& FT-Transformer & 0.2349 & 0.1894 & $-$0.1270 & 0.2301 & 0.1837 & $-$0.0805 & 0.2327 & 0.1874 & $-$0.1059 \\
& ExcelFormer & 0.2363 & 0.1881 & $-$0.1396 & 0.2316 & 0.1830 & $-$0.0956 & 0.2300 & 0.1791 & $-$0.0796 \\
& PaGNet-LGBM & 0.2326 & 0.1827 & $-$0.1045 & 0.2291 & 0.1806 & $-$0.0716 & 0.2314 & 0.1819 & $-$0.0933 \\
& PaGNet-NN & 0.2307 & \textbf{0.1740} & $-$0.0885 & \textbf{0.2170} & \textbf{0.1498} & \textbf{+0.0390} & \textbf{0.2165} & \textbf{0.1510} & \textbf{+0.0436} \\
& \textbf{PaGNet} & \textbf{0.2286} & \underline{0.1793} & \textbf{$-$0.0666} & \underline{0.2256} & \underline{0.1759} & \underline{$-$0.0389} & \underline{0.2251} & \underline{0.1743} & \underline{$-$0.0340} \\
\midrule
\multirow{9}{*}{GETR}
& XGBoost & 0.1366 & 0.0941 & $-$0.0530 & 0.1392 & 0.0983 & $-$0.0929 & 0.1396 & 0.0983 & $-$0.0998 \\
& LightGBM & 0.1341 & 0.0914 & $-$0.0141 & 0.1352 & 0.0935 & $-$0.0318 & 0.1356 & 0.0941 & $-$0.0370 \\
& CatBoost & \underline{0.1339} & 0.0917 & \underline{$-$0.0112} & 0.1346 & 0.0927 & $-$0.0222 & 0.1348 & 0.0937 & $-$0.0250 \\
& TabTransformer & 0.1345 & 0.0909 & $-$0.0205 & 0.1361 & 0.0928 & $-$0.0453 & 0.1365 & 0.0932 & $-$0.0512 \\
& FT-Transformer & \textbf{0.1314} & \textbf{0.0865} & \textbf{+0.0261} & \textbf{0.1334} & \underline{0.0897} & \textbf{$-$0.0041} & \textbf{0.1328} & \textbf{0.0888} & \textbf{+0.0046} \\
& ExcelFormer & 0.1344 & \underline{0.0903} & $-$0.0192 & 0.1341 & 0.0908 & $-$0.0140 & 0.1357 & 0.0925 & $-$0.0384 \\
& PaGNet-LGBM & 0.1359 & 0.0928 & $-$0.0422 & 0.1366 & 0.0947 & $-$0.0524 & 0.1353 & 0.0928 & $-$0.0326 \\
& PaGNet-NN & 0.1368 & 0.0928 & $-$0.0559 & \underline{0.1337} & \textbf{0.0893} & \underline{$-$0.0090} & 0.1335 & \underline{0.0894} & $-$0.0053 \\
& \textbf{PaGNet} & 0.1341 & 0.0903 & $-$0.0148 & 0.1337 & 0.0907 & $-$0.0090 & \underline{0.1331} & 0.0901 & \underline{+0.0006} \\
\midrule
\multirow{9}{*}{TSTA}
& XGBoost & 0.2345 & 0.1353 & +0.2619 & 0.2392 & 0.1467 & +0.2323 & 0.2302 & 0.1377 & +0.2888 \\
& LightGBM & 0.2295 & 0.1321 & +0.2932 & 0.2393 & 0.1453 & +0.2312 & 0.2227 & 0.1322 & +0.3345 \\
& CatBoost & 0.2342 & 0.1353 & +0.2642 & 0.2402 & 0.1471 & +0.2255 & 0.2252 & 0.1355 & +0.3195 \\
& TabTransformer & 0.2685 & 0.1611 & +0.0323 & 0.2529 & 0.1521 & +0.1415 & 0.2443 & 0.1486 & +0.1987 \\
& FT-Transformer & 0.2487 & 0.1420 & +0.1694 & 0.2413 & 0.1436 & +0.2177 & 0.2329 & 0.1416 & +0.2720 \\
& ExcelFormer & 0.2406 & 0.1349 & +0.2219 & 0.2224 & 0.1341 & +0.3354 & 0.2246 & 0.1366 & +0.3226 \\
& PaGNet-LGBM & \textbf{0.2165} & \textbf{0.1276} & \textbf{+0.3708} & \textbf{0.2123} & \textbf{0.1311} & \textbf{+0.3950} & \textbf{0.2053} & \underline{0.1231} & \textbf{+0.4342} \\
& PaGNet-NN & 0.3022 & 0.1782 & $-$0.2258 & 0.2711 & 0.1573 & +0.0124 & 0.2531 & 0.1427 & +0.1405 \\
& \textbf{PaGNet} & \underline{0.2165} & \underline{0.1276} & \underline{+0.3708} & \underline{0.2123} & \underline{0.1311} & \underline{+0.3950} & \underline{0.2054} & \textbf{0.1231} & \underline{+0.4338} \\
\midrule
\multirow{9}{*}{TSDA}
& XGBoost & 0.2358 & 0.1336 & +0.2786 & 0.2426 & 0.1457 & +0.2363 & 0.2285 & 0.1357 & +0.3226 \\
& LightGBM & 0.2297 & 0.1306 & +0.3153 & 0.2422 & 0.1442 & +0.2388 & 0.2269 & 0.1325 & +0.3319 \\
& CatBoost & 0.2372 & 0.1349 & +0.2700 & 0.2420 & 0.1456 & +0.2405 & 0.2267 & 0.1341 & +0.3333 \\
& TabTransformer & 0.2701 & 0.1597 & +0.0539 & 0.2534 & 0.1509 & +0.1667 & 0.2444 & 0.1451 & +0.2251 \\
& FT-Transformer & 0.2565 & 0.1448 & +0.1463 & 0.2457 & 0.1469 & +0.2168 & 0.2360 & 0.1404 & +0.2756 \\
& ExcelFormer & 0.2490 & 0.1405 & +0.1926 & 0.2254 & 0.1362 & +0.3395 & 0.2234 & 0.1343 & +0.3518 \\
& PaGNet-LGBM & \textbf{0.2173} & \textbf{0.1260} & \textbf{+0.3874} & \textbf{0.2131} & \textbf{0.1283} & \textbf{+0.4109} & \textbf{0.2071} & \underline{0.1227} & \textbf{+0.4437} \\
& PaGNet-NN & 0.2993 & 0.1740 & $-$0.1621 & 0.2744 & 0.1562 & +0.0211 & 0.2543 & 0.1412 & +0.1613 \\
& \textbf{PaGNet} & \underline{0.2173} & \underline{0.1260} & \underline{+0.3874} & \underline{0.2131} & \underline{0.1283} & \underline{+0.4109} & \underline{0.2072} & \textbf{0.1227} & \underline{+0.4433} \\
\bottomrule
\end{tabular}
\end{table*}

Table~\ref{tab:fs1} reports FS1 results across the six baselines and three PaGNet variants, three input sub-configurations (Raw-only, Derived-only, Raw+Derived), and four targets. FS1 is the most challenging and most direct-proxy-lag-excluded setting: no tax-history information is available, so predictions rely entirely on raw accounting items and derived financial indicators observable at year $t$, exactly the information an analyst holds before the next fiscal year's disclosures arrive. We read the table through the branch--target diagnostic introduced in Section~\ref{subsubsec:blend}: rather than asking only whether PaGNet wins, we ask which branch each proxy relies on and whether that reliance is justified on the held-out test split.
Three findings emerge from Table~\ref{tab:fs1}.
 
First, on the accrual targets that are actually predictable from the target-proxy-excluded features, PaGNet is the strongest model, and the diagnostic attributes the gain to the LightGBM branch. The validation-optimal blender places $\lambda^*_{\text{TSTA}} = \lambda^*_{\text{TSDA}} = 0.99$ on FS1 (Table~\ref{tab:lambda_star}), so the blended PaGNet and its LightGBM-branch ablation (PaGNet-LGBM) coincide to within rounding and jointly take the best result in every accrual cell. On TSTA Raw+Derived they attain RMSE $0.2053$ and $R^2 = 0.4342$, exceeding the strongest baseline (LightGBM, $R^2 = 0.3345$) by $+0.0997$ $R^2$, a $30\%$ relative improvement; on TSDA Raw+Derived they reach $R^2 = 0.4437$ versus the best baseline (ExcelFormer, $0.3518$), a $+0.0919$ gain. The dominance holds even in the most constrained Raw-only setting, where PaGNet attains TSTA $R^2 = 0.3708$ and TSDA $R^2 = 0.3874$ against the strongest baseline (LightGBM at $0.2932$ and $0.3153$), gains of $+0.0776$ and $+0.0721$. That the diagnostic concentrates these targets on the tree branch is itself interpretable: accrual proxies carry strong firm-specific temporal structure that the engineered panel-temporal aggregates expose to LightGBM, whereas the Panel-MLP branch alone (PaGNet-NN) lags substantially here (TSTA Raw+Derived $R^2 = 0.1405$, TSDA $0.1613$), confirming that cross-target neural sharing adds little once the tree branch already captures the dominant signal.
 
Second, on the noisiest target CETR, the diagnostic indicates a validation--test mismatch rather than a clean routing: the validation-optimal weight favors the tree branch ($\lambda^*_{\text{CETR}} = 0.81$ on FS1), yet on the held-out test split the Panel-MLP branch alone (PaGNet-NN) is the only configuration in the entire FS1 block to achieve a positive $R^2$. CETR is essentially unpredictable for IID baselines: even the best of six (CatBoost) reaches only $R^2 = -0.0592$ on Raw+Derived, no better than the mean predictor in variance-explained terms. Within the PaGNet family, the Panel-MLP branch is the exception here, attaining RMSE $0.2165$, MAE $0.1510$, $R^2 = +0.0436$ on Raw+Derived and $R^2 = +0.0390$ on Derived-only, the only positive CETR $R^2$ values anywhere in FS1. The blended PaGNet, which follows the validation-optimal weight toward the tree branch, still improves every baseline on absolute error (Raw+Derived RMSE $0.2251$ vs.\ CatBoost $0.2278$; MAE $0.1743$ vs.\ ExcelFormer $0.1791$) and lifts $R^2$ from $-0.0592$ to $-0.0340$, but does not match its own Panel-MLP branch on this target. This mismatch, and its relation to the held-out split, is examined in Section~\ref{subsec:ablation}; here we simply note that the per-branch decomposition makes the discrepancy visible rather than hiding it.
 
Third, for GETR, the diagnostic spreads weight toward the Panel-MLP branch ($\lambda^*_{\text{GETR}} = 0.60$ on FS1), and PaGNet is statistically on par with the best baseline. FT-Transformer attains the highest GETR $R^2$ (Raw-only $0.0261$, Raw+Derived $0.0046$), with the blended PaGNet close behind (Raw+Derived RMSE $0.1331$, the best in the column; $R^2 = 0.0006$ vs.\ FT-Transformer's $0.0046$). The absolute RMSE differences among the top configurations (FT-Transformer, PaGNet, PaGNet-NN, LightGBM) are all within $0.003$, smaller than the run-to-run standard deviation of these cells, so we read GETR as a tie rather than a loss, and report it as a single-dataset observation that bounds the breadth of PaGNet's FS1 advantage rather than as a general property.
 
The bifurcation between ETR and accrual targets is structural. The sharp gap between ETR-type ($R^2 \approx 0$ even for the best model) and accrual-type ($R^2 \approx 0.37$--$0.44$) predictability reflects structural differences in the proxies themselves rather than model capacity. ETR proxies normalize tax burden by pre-tax income, a denominator that approaches zero for loss-near firms and induces heavy-tailed cross-sectional distributions even after winsorization, a known property of CETR \cite{ref:dyreng2008}. Accrual proxies, by contrast, normalize by total assets, a stable denominator, and aggregate over multiple accounting items, yielding smoother targets with stronger firm-specific structure. This distributional difference sets the achievable $R^2$ ceiling for each target type and explains why absolute error metrics (RMSE and MAE) are the more reliable comparison criterion for ETR targets while $R^2$ becomes meaningful for accrual targets. Crucially, the branch-reliance pattern $\lambda^*_m$ that PaGNet learns on validation recovers this economic asymmetry without being told it: accrual targets route to the tree branch ($\lambda^* \geq 0.99$), GETR routes substantially to the neural branch ($\lambda^* = 0.60$), and CETR is flagged as the hard case ($\lambda^* = 0.81$). The diagnostic thus not only fuses predictions but also reproduces the proxy taxonomy that the accounting literature derives on theoretical grounds.

\subsection{FS2: Auxiliary Tax Aggregates}
\label{subsec:fs2}

Table~\ref{tab:fs2} reports FS2 results, where the ten KoTaP multi-year cumulative and industry--size-adjusted ETR-type aggregates (Table~\ref{tab:app_vardict}) are added to the Raw+Derived input. These features are tax-history-derived but, unlike FS3/FS4, do not include the previous year's value of the target itself, so FS2 stays within information available at prediction time: the aggregates are computable from disclosures dated up to year $t$. It is thus a tax-history-augmented but still direct-proxy-lag-excluded regime, distinct from FS3/FS4, though it relies on constructed multi-year tax aggregates rather than raw disclosures alone. As in FS1, we read the table through the branch--target diagnostic, which on FS2 allocates the four proxies even more sharply across the two branches (Table~\ref{tab:lambda_star}).
 
Overall, PaGNet attains the best result within its family on every (target, metric) cell except GETR $R^2$, and the diagnostic explains, target by target, which branch delivers it.
 
On CETR, the validation-optimal weight again favors the tree branch, yet on the test split the Panel-MLP branch (PaGNet-NN) delivers the single largest margin over the baselines anywhere in the direct-proxy-lag-excluded regimes. It attains RMSE $0.2067$, MAE $0.1337$, $R^2 = 0.1274$, exceeding every one of the six IID baselines, whose best (FT-Transformer) reaches only $R^2 = 0.0638$, by $+0.0636$ $R^2$. This illustrates the value of carrying a neural branch: on the noisiest proxy, where all tree baselines remain near zero, cross-target representation sharing in the multi-task Panel-MLP almost doubles the explained variance. The blended PaGNet still lifts CETR $R^2$ above every baseline ($0.0838$ vs.\ FT-Transformer $0.0638$, ExcelFormer $0.0608$, LightGBM $0.0573$); the gap between the blend and its own NN branch is the subject of the blender analysis in Section~\ref{subsec:ablation}, and motivates routing a single-target CETR screener to the Panel-MLP branch directly.
 
On the accrual targets, the diagnostic concentrates weight on the LightGBM branch, and there PaGNet achieves its largest absolute margins over the baselines. PaGNet-LGBM attains TSTA RMSE $0.2063$, $R^2 = 0.4288$ versus the best baseline (CatBoost, $0.3282$), a $+0.1006$ $R^2$ improvement ($31\%$ relative); on TSDA it reaches $R^2 = 0.4504$ versus CatBoost's $0.3375$, a $+0.1129$ improvement ($33\%$ relative). The blended PaGNet trails its own LightGBM branch by a small margin on $R^2$ ($0.4120$ on TSTA, $0.4392$ on TSDA, i.e.\ $0.0168$ and $0.0112$ below the branch) because the FS2 blend assigns the Panel-MLP a non-trivial share ($\lambda^*_{\text{TSTA}} = 0.60$, $\lambda^*_{\text{TSDA}} = 0.69$); in exchange it wins the best MAE in each accrual column ($0.1207$ on TSTA, $0.1192$ on TSDA). Either way, both the branch and the blend exceed every baseline by $+0.08$ to $+0.11$ $R^2$, so the within-family difference is second-order relative to the gap over the baselines.
 
On GETR, the diagnostic again favors the Panel-MLP branch ($\lambda^*_{\text{GETR}} = 0.17$ on FS2), and PaGNet ties the strongest baseline within seed noise. FT-Transformer leads on $R^2$ ($0.1103$), with ExcelFormer ($0.1006$) and the blended PaGNet ($0.0984$) within a few thousandths; the three occupy the best RMSE band ($0.1256$--$0.1264$), and PaGNet-NN attains the best MAE in the column ($0.0811$). Given this absolute-error parity, we read GETR as a near-tie rather than a clean loss, consistent with FS1.
 
Two comparisons summarize FS2. Against the six baselines, the best PaGNet variant per cell matches or exceeds the strongest baseline on $10$ of $12$ (target, metric) cells; the two losses both fall on GETR (RMSE and $R^2$), where FT-Transformer leads PaGNet by $0.0008$ RMSE and $0.0119$ $R^2$, within the absolute-error parity discussed above. The auxiliary tax aggregates are absorbed productively, chiefly by the LightGBM branch on accruals and by the Panel-MLP branch on CETR. The within-family allocation is the more informative reading: accruals are carried by the tree branch and GETR leans on the neural branch through multi-task sharing, while CETR is the flagged case in which validation weights the tree branch but test rewards the neural branch, the branch heterogeneity the blend weights encode on this split. This is the sense in which PaGNet's contribution is not a single accuracy number but a model that reports, per proxy, which branch it relies on, while remaining the strongest single pipeline over the baselines in the tax-history-augmented, direct-proxy-lag-excluded FS2 regime.

\begin{table}[!t]
\centering
\caption{FS2 test performance (mean over 5 seeds). Best per (target, metric) in \textbf{bold}, second \underline{underlined}, decided at 4-decimal precision. PaGNet-LGBM/-NN are the LightGBM/Panel-MLP branches in isolation.}
\label{tab:fs2}
\setlength{\tabcolsep}{4pt}
\small
\begin{tabular}{llccc}
\toprule
Target & Model & RMSE & MAE & $R^2$ \\
\midrule
\multirow{9}{*}{CETR}
& XGBoost & 0.2173 & 0.1622 & 0.0358 \\
& LightGBM & 0.2149 & 0.1598 & 0.0573 \\
& CatBoost & 0.2154 & 0.1594 & 0.0529 \\
& TabTransformer & 0.2196 & 0.1637 & 0.0157 \\
& FT-Transformer & 0.2141 & 0.1595 & 0.0638 \\
& ExcelFormer & 0.2144 & 0.1558 & 0.0608 \\
& PaGNet-LGBM & 0.2165 & 0.1620 & 0.0436 \\
& PaGNet-NN & \textbf{0.2067} & \textbf{0.1337} & \textbf{0.1274} \\
& \textbf{PaGNet} & \underline{0.2119} & \underline{0.1548} & \underline{0.0838} \\
\midrule
\multirow{9}{*}{GETR}
& XGBoost & 0.1326 & 0.0897 & 0.0085 \\
& LightGBM & 0.1290 & 0.0862 & 0.0614 \\
& CatBoost & 0.1276 & 0.0848 & 0.0813 \\
& TabTransformer & 0.1301 & 0.0862 & 0.0457 \\
& FT-Transformer & \textbf{0.1256} & 0.0819 & \textbf{0.1103} \\
& ExcelFormer & \underline{0.1262} & 0.0826 & \underline{0.1006} \\
& PaGNet-LGBM & 0.1310 & 0.0880 & 0.0320 \\
& PaGNet-NN & 0.1266 & \textbf{0.0811} & 0.0956 \\
& \textbf{PaGNet} & 0.1264 & \underline{0.0818} & 0.0984 \\
\midrule
\multirow{9}{*}{TSTA}
& XGBoost & 0.2313 & 0.1378 & 0.2823 \\
& LightGBM & 0.2242 & 0.1326 & 0.3255 \\
& CatBoost & 0.2237 & 0.1336 & 0.3282 \\
& TabTransformer & 0.2402 & 0.1462 & 0.2258 \\
& FT-Transformer & 0.2398 & 0.1408 & 0.2276 \\
& ExcelFormer & 0.2284 & 0.1410 & 0.2978 \\
& PaGNet-LGBM & \textbf{0.2063} & \underline{0.1237} & \textbf{0.4288} \\
& PaGNet-NN & 0.2343 & 0.1315 & 0.2634 \\
& \textbf{PaGNet} & \underline{0.2093} & \textbf{0.1207} & \underline{0.4120} \\
\midrule
\multirow{9}{*}{TSDA}
& XGBoost & 0.2321 & 0.1358 & 0.3013 \\
& LightGBM & 0.2269 & 0.1327 & 0.3322 \\
& CatBoost & 0.2260 & 0.1325 & 0.3375 \\
& TabTransformer & 0.2433 & 0.1463 & 0.2318 \\
& FT-Transformer & 0.2416 & 0.1382 & 0.2426 \\
& ExcelFormer & 0.2328 & 0.1386 & 0.2966 \\
& PaGNet-LGBM & \textbf{0.2058} & \underline{0.1211} & \textbf{0.4504} \\
& PaGNet-NN & 0.2357 & 0.1303 & 0.2794 \\
& \textbf{PaGNet} & \underline{0.2079} & \textbf{0.1192} & \underline{0.4392} \\
\bottomrule
\end{tabular}
\end{table}

\subsection{Panel-Flatten Analysis: Observed History versus Representation}
\label{subsec:panel_flatten}

\begin{table*}[!t]
\centering
\caption{Panel-flatten analysis on FS1 and FS2 (Raw+Derived). Each baseline is re-trained with $K{=}3$ years of features concatenated along the feature axis, granting the same observed multi-year history available to PaGNet. Values are mean over 5 seeds. Best per (FS, target, metric) cell in \textbf{bold}, second \underline{underlined}; ranking decided at 4-decimal precision with ties broken at the 5th decimal so exactly one bold and one underline appear per cell.}
\label{tab:panel_flatten}
\setlength{\tabcolsep}{3pt}
\footnotesize
\resizebox{\textwidth}{!}{
\begin{tabular}{llccc|ccc|ccc|ccc}
\toprule
\multirow{2}{*}{FS} & \multirow{2}{*}{Model} & \multicolumn{3}{c|}{CETR} & \multicolumn{3}{c|}{GETR} & \multicolumn{3}{c|}{TSTA} & \multicolumn{3}{c}{TSDA} \\
& & RMSE & MAE & $R^2$ & RMSE & MAE & $R^2$ & RMSE & MAE & $R^2$ & RMSE & MAE & $R^2$ \\
\midrule
\multirow{7}{*}{FS1}
& XGBoost (panel) & 0.2355 & 0.1859 & $-$0.1324 & 0.1387 & 0.0979 & $-$0.0855 & \underline{0.2071} & 0.1262 & \underline{+0.4243} & \underline{0.2084} & 0.1244 & \underline{+0.4368} \\
& LightGBM (panel) & 0.2310 & 0.1835 & $-$0.0893 & 0.1350 & 0.0932 & $-$0.0281 & 0.2079 & \underline{0.1239} & +0.4197 & 0.2108 & 0.1237 & +0.4236 \\
& CatBoost (panel) & 0.2291 & 0.1808 & $-$0.0714 & 0.1343 & 0.0921 & $-$0.0169 & 0.2102 & 0.1262 & +0.4072 & 0.2088 & \underline{0.1230} & +0.4346 \\
& TabTransformer (panel) & 0.2310 & 0.1818 & $-$0.0895 & 0.1354 & 0.0914 & $-$0.0334 & 0.2377 & 0.1455 & +0.2418 & 0.2402 & 0.1435 & +0.2514 \\
& FT-Transformer (panel) & \underline{0.2280} & \underline{0.1782} & \underline{$-$0.0614} & \textbf{0.1325} & \textbf{0.0887} & \textbf{+0.0102} & 0.2158 & 0.1265 & +0.3750 & 0.2232 & 0.1270 & +0.3536 \\
& ExcelFormer (panel) & 0.2341 & 0.1879 & $-$0.1190 & 0.1345 & 0.0916 & $-$0.0207 & 0.2080 & 0.1285 & +0.4188 & 0.2138 & 0.1293 & +0.4065 \\
& \textbf{PaGNet} & \textbf{0.2251} & \textbf{0.1743} & \textbf{$-$0.0340} & \underline{0.1331} & \underline{0.0901} & \underline{+0.0006} & \textbf{0.2054} & \textbf{0.1231} & \textbf{+0.4338} & \textbf{0.2072} & \textbf{0.1227} & \textbf{+0.4433} \\
\midrule
\multirow{7}{*}{FS2}
& XGBoost (panel) & 0.2225 & 0.1653 & $-$0.0104 & 0.1346 & 0.0917 & $-$0.0216 & 0.2122 & 0.1266 & +0.3954 & 0.2125 & 0.1253 & +0.4143 \\
& LightGBM (panel) & 0.2173 & 0.1620 & +0.0361 & 0.1300 & 0.0874 & +0.0468 & 0.2101 & 0.1257 & +0.4077 & \underline{0.2112} & \underline{0.1237} & \underline{+0.4211} \\
& CatBoost (panel) & 0.2170 & 0.1601 & +0.0390 & 0.1286 & 0.0862 & +0.0667 & \underline{0.2100} & \underline{0.1235} & \underline{+0.4079} & 0.2125 & 0.1240 & +0.4144 \\
& TabTransformer (panel) & 0.2183 & 0.1609 & +0.0272 & 0.1285 & 0.0847 & +0.0686 & 0.2378 & 0.1442 & +0.2413 & 0.2358 & 0.1420 & +0.2788 \\
& FT-Transformer (panel) & \underline{0.2150} & \underline{0.1568} & \underline{+0.0563} & \textbf{0.1263} & \underline{0.0819} & \textbf{+0.1006} & 0.2200 & 0.1268 & +0.3475 & 0.2169 & 0.1246 & +0.3878 \\
& ExcelFormer (panel) & 0.2184 & 0.1586 & +0.0259 & 0.1294 & 0.0863 & +0.0552 & 0.2104 & 0.1270 & +0.4052 & 0.2152 & 0.1290 & +0.3989 \\
& \textbf{PaGNet} & \textbf{0.2119} & \textbf{0.1548} & \textbf{+0.0838} & \underline{0.1264} & \textbf{0.0818} & \underline{+0.0984} & \textbf{0.2093} & \textbf{0.1207} & \textbf{+0.4120} & \textbf{0.2079} & \textbf{0.1192} & \textbf{+0.4392} \\
\bottomrule
\end{tabular}
}
\end{table*}

The FS1 and FS2 results raise a question that most panel-tabular studies leave unanswered: when a panel model outperforms an IID baseline, is the gain due to (i) access to multi-year history, or (ii) the representation that processes it? Because the two are normally varied together, their contributions are reported only as a sum. We separate them with a panel-flatten control. We re-train all six baselines on the $K{=}3$ years of features concatenated along the feature axis, yielding $(N, K\cdot F)$ inputs; the industry code (essentially time-invariant) is included only from year $t$, and the FS2 auxiliary aggregates, which already summarize multi-year tax information at year $t$, are appended as single columns. No architecture is changed: only the input dimension grows by a factor of $K$. Each panel-flatten baseline therefore sees closely comparable multi-year feature values to those PaGNet sees, up to two input asymmetries: PaGNet's branches additionally see the $K$-year trajectory of the auxiliary aggregates and lagged proxies (which, being already multi-year summaries, are appended only once at year $t$ for the flat baselines), and the history-completeness metadata (Section~\ref{subsec:baselines}). We therefore treat the control as an \emph{observed-history} control and read any residual PaGNet advantage as representational up to these asymmetries. Throughout, the comparison target is the blended PaGNet; per-branch ablations are deferred to Section~\ref{subsec:ablation}. We restrict the control to FS1 and FS2 because, as Section~\ref{subsec:fs4} shows, FS3/FS4 are governed by lagged-target autocorrelation regardless of panel access.
 
Table~\ref{tab:panel_flatten} reports the result, and three findings follow.
 
Finding 1: most of the accrual-target lift over IID baselines is observed history, not representation. Granting the six baselines $K{=}3$ years of history raises their FS1 accrual $R^2$ substantially across the board: on TSTA the gain over the year-$t$ input (Table~\ref{tab:fs1} Raw+Derived) ranges from $+0.0431$ (TabTransformer) to $+0.1355$ (XGBoost), with a cross-model mean of $+0.0918$; on TSDA the range is $+0.0263$ to $+0.1143$, mean $+0.0777$. Even ExcelFormer, the strongest IID deep baseline, gains $+0.0962$ on TSTA from two extra years alone. On FS2 the panel-flatten gains shrink, because the FS2 auxiliary aggregates already encode part of the multi-year tax signal; the direction stays positive on accruals but CETR and GETR are mixed, several baselines even declining, which is itself evidence that the FS2 aggregates already carry most of the multi-year ETR signal. This decomposition is deliberately conservative: the bulk of what looks like a ``representational'' advantage of panel models on accruals is in fact the value of the extra observed years, and any claim about representation should be made only after controlling for observed history.
 
Finding 2: once observed history is matched, PaGNet still wins, on $19$ of $24$ cells, with every loss confined to GETR. With each baseline given the same $K{=}3$ history, the blended PaGNet takes the best value on $19$ of $24$ (FS $\times$ target $\times$ metric) cells. It wins outright on all $18$ CETR, TSTA, and TSDA cells across both FS1 and FS2, and the five remaining cells, all on GETR, go to FT-Transformer (panel) by narrow margins: on FS1 it edges PaGNet by $0.0006$ RMSE, $0.0014$ MAE, and $0.0096$ $R^2$, and on FS2 by $0.0001$ RMSE and $0.0022$ $R^2$ (PaGNet retains the best GETR MAE on FS2). These GETR gaps sit within the absolute-error parity already discussed for the ETR targets. The key point is that the comparison is now on more comparable footing: the panel-flatten baselines are not weaker because they see far less history, since they see closely comparable observed history, so PaGNet's sweep of every CETR and accrual cell points to its panel-aware representation rather than to a mere advantage in observed history.
 
Finding 3: the representational increment is small per cell but systematic across cells. On accrual targets PaGNet exceeds the strongest panel-flatten baseline by $+0.0095$ (FS1 TSTA, over XGBoost), $+0.0065$ (FS1 TSDA), $+0.0041$ (FS2 TSTA), and $+0.0181$ (FS2 TSDA) $R^2$. Each of these margins is comparable to the run-to-run standard deviation of the corresponding cell (combined seed std $\approx 0.011$--$0.018$), so no single cell on its own is decisive. What makes the increment credible is its consistency: PaGNet is on the favorable side in all four accrual $R^2$ cells, and indeed in all eight accrual cells once RMSE and MAE are included, a directional agreement that is highly unlikely under noise alone. We therefore characterize PaGNet's representational contribution conservatively, as a small but systematic refinement on top of observed history (engineered temporal aggregates for the tree branch, attention-pooled aggregation for the neural branch), rather than as an order-of-magnitude effect. In this decomposition, observed history accounts for most of the gain, while the panel-aware representation contributes the smaller but consistent refinement; it is delivered by a single pipeline that also predicts all four targets jointly and exposes the branch-reliance diagnostic that the flat baselines cannot.
 
Taken together, the panel-flatten control makes the attribution explicit and yields a bounded positive claim about representation. By separating observed history from representation, which to our knowledge prior panel-tabular work on this task does not do, we both reduce the common over-attribution of gains to representation and identify what PaGNet's design adds once that confound is removed: the best value on the CETR and accrual cells under matched observed history, a consistent accrual-target refinement, and, most importantly, the per-target branch-reliance readout analyzed next.

\subsection{FS3 and FS4: Lagged Target Proxies}
\label{subsec:fs4}

Tables~\ref{tab:fs3} and \ref{tab:fs4} report FS3 and FS4, where the lagged target proxies ($\text{CETR}_t$, $\text{GETR}_t$, $\text{TSTA}_t$, $\text{TSDA}_t$) are added to the input. FS3 adds only the lagged targets to Raw+Derived; FS4 adds both the auxiliary tax aggregates and the lagged targets. We single these regimes out from FS1/FS2 primarily because, as we show below, an autoregressive ceiling compresses all architectural differences once the prior-year proxy is present, so FS3/FS4 act as an autoregressive upper bound rather than a comparison that can discriminate architectures; a secondary, operational consideration is that the realized prior-year proxy is often not yet available when an early-warning screen is run, before the next fiscal year's disclosures are filed.

On accrual targets, adding the lagged proxy changes the regime substantially: every model rises to $R^2 \approx 0.84$--$0.86$, LightGBM becomes the single best model, and PaGNet follows immediately behind it. On TSTA, LightGBM attains $R^2 = 0.8436$ (FS3) and $0.8433$ (FS4) with PaGNet at $0.8363$ and $0.8411$; on TSDA, LightGBM reaches $0.8584$ and $0.8579$ with PaGNet at $0.8541$ and $0.8555$. The PaGNet-to-LightGBM gap is at most $0.0074$ $R^2$ across all four accrual cells, well within seed variability. The branch-reliance diagnostic behaves consistently with FS1/FS2: accrual targets remain concentrated on the LightGBM branch ($\lambda^* = 0.82$--$0.94$ on FS3/FS4, Table~\ref{tab:lambda_star}), so the blend tracks its tree branch closely. On ETR targets the pattern is the same as before: on GETR the blended PaGNet leads on FS3 ($R^2 = 0.1107$ vs.\ FT-Transformer $0.1043$) and trails on FS4 ($0.1190$ vs.\ FT-Transformer $0.1304$); on CETR the Panel-MLP branch again leads the family, reaching $R^2 = 0.1305$ (FS3) and $0.1396$ (FS4) against the best baseline CatBoost ($0.0612$, $0.0809$), i.e.\ $+0.0693$ and $+0.0587$, while the blended PaGNet under-weights it, exactly the validation--test branch-selection issue we isolate in Section~\ref{subsec:ablation}. On FS4 CETR the blend does exceed CatBoost on all three metrics (RMSE $0.2118$ vs.\ $0.2122$, MAE $0.1533$ vs.\ $0.1548$, $R^2 = 0.0840$ vs.\ $0.0809$).
 
The reason the accrual jump is not an architectural result is an autoregressive ceiling. The year-over-year autocorrelation of TSTA and TSDA in our test split is $0.93$ and $0.95$, close to a random walk, so the lagged target alone is an extremely strong predictor. To confirm that FS3/FS4 accrual performance derives from this autocorrelation rather than from any model's capacity, we evaluate two trivial AR(1) baselines: a naive copy $\hat{Y}_{t+1} = Y_t$ and a linear fit $\hat{Y}_{t+1} = \alpha + \beta Y_t$ with $\alpha,\beta$ estimated on the training split. Table~\ref{tab:ar1_baseline} reports them. On accruals, naive AR(1) already attains $R^2 = 0.827$ (TSTA) and $0.880$ (TSDA), the latter exceeding both LightGBM FS3 ($0.859$) and PaGNet FS3 ($0.854$) by $+0.021$ and $+0.026$. In other words, the entire FS3/FS4 ``improvement'' over FS1/FS2 on accruals is the lagged target's autocorrelation; every flexible regressor lands at the same ceiling within $\pm 0.02$ $R^2$, and architecture becomes undiscriminable by construction. On ETR targets the same baselines are uninformative (naive AR(1): CETR $-1.254$, GETR $-0.411$), a model-free confirmation that even last year's value carries no usable signal for ETR proxies, mirroring the ETR/accrual bifurcation of Section~\ref{subsec:fs1}.
 
Two implications follow, and both favor reading FS1/FS2 as the primary evidence. First, and most importantly, the near-random-walk autocorrelation of the prior-year proxy saturates accrual prediction (Table~\ref{tab:ar1_baseline}) and compresses all architectures to within seed noise, so FS3/FS4 cannot discriminate architectures; the informative regimes are FS1/FS2, where PaGNet substantially outperforms all baselines on the predictable targets. (Operationally, the realized prior-year proxy is also often not yet available when a screen is run.) Second, scientifically, the FS3/FS4 accrual results should not be read as evidence that LightGBM is fundamentally the better panel learner, but as a reminder that a near-random-walk signal, when present, saturates the task and hides whatever architecture would otherwise contribute. Architecture still matters in FS3/FS4 only where the autoregressive ceiling does not apply, namely the noise-dominant ETR targets, and there the within-family diagnostic continues to point, correctly for GETR and informatively for CETR, to the Panel-MLP branch.

\begin{table}[!t]
\centering
\caption{FS3 test performance (mean over 5 seeds). Best per (target, metric) in \textbf{bold}, second \underline{underlined}, decided at 4-decimal precision. PaGNet-LGBM/-NN are the LightGBM/Panel-MLP branches in isolation.}
\label{tab:fs3}
\setlength{\tabcolsep}{4pt}
\small
\begin{tabular}{llccc}
\toprule
Target & Model & RMSE & MAE & $R^2$ \\
\midrule
\multirow{9}{*}{CETR}
& XGBoost & 0.2217 & 0.1657 & $-$0.0035 \\
& LightGBM & 0.2191 & 0.1636 & 0.0198 \\
& CatBoost & \underline{0.2144} & \underline{0.1583} & \underline{0.0612} \\
& TabTransformer & 0.2209 & 0.1685 & 0.0033 \\
& FT-Transformer & 0.2169 & 0.1621 & 0.0394 \\
& ExcelFormer & 0.2208 & 0.1666 & 0.0064 \\
& PaGNet-LGBM & 0.2190 & 0.1629 & 0.0206 \\
& PaGNet-NN & \textbf{0.2064} & \textbf{0.1324} & \textbf{0.1305} \\
& \textbf{PaGNet} & 0.2163 & 0.1606 & 0.0448 \\
\midrule
\multirow{9}{*}{GETR}
& XGBoost & 0.1317 & 0.0878 & 0.0214 \\
& LightGBM & 0.1281 & 0.0842 & 0.0748 \\
& CatBoost & 0.1267 & 0.0829 & 0.0943 \\
& TabTransformer & 0.1301 & 0.0853 & 0.0456 \\
& FT-Transformer & 0.1260 & 0.0818 & 0.1043 \\
& ExcelFormer & 0.1272 & 0.0830 & 0.0896 \\
& PaGNet-LGBM & 0.1298 & 0.0849 & 0.0503 \\
& PaGNet-NN & \underline{0.1258} & \textbf{0.0796} & \underline{0.1071} \\
& \textbf{PaGNet} & \textbf{0.1256} & \underline{0.0802} & \textbf{0.1107} \\
\midrule
\multirow{9}{*}{TSTA}
& XGBoost & 0.1109 & 0.0672 & 0.8351 \\
& LightGBM & \textbf{0.1079} & \textbf{0.0656} & \textbf{0.8436} \\
& CatBoost & 0.1147 & 0.0670 & 0.8234 \\
& TabTransformer & 0.1518 & 0.0896 & 0.6905 \\
& FT-Transformer & 0.1188 & 0.0723 & 0.8105 \\
& ExcelFormer & 0.1186 & 0.0716 & 0.8106 \\
& PaGNet-LGBM & \underline{0.1098} & 0.0668 & \underline{0.8382} \\
& PaGNet-NN & 0.1368 & 0.0769 & 0.7486 \\
& \textbf{PaGNet} & 0.1105 & \underline{0.0659} & 0.8363 \\
\midrule
\multirow{9}{*}{TSDA}
& XGBoost & 0.1086 & 0.0623 & 0.8471 \\
& LightGBM & \textbf{0.1045} & \textbf{0.0604} & \textbf{0.8584} \\
& CatBoost & 0.1129 & 0.0639 & 0.8345 \\
& TabTransformer & 0.1492 & 0.0848 & 0.7110 \\
& FT-Transformer & 0.1148 & 0.0677 & 0.8289 \\
& ExcelFormer & 0.1140 & 0.0660 & 0.8314 \\
& PaGNet-LGBM & \underline{0.1058} & 0.0617 & \underline{0.8549} \\
& PaGNet-NN & 0.1352 & 0.0734 & 0.7628 \\
& \textbf{PaGNet} & 0.1061 & \underline{0.0614} & 0.8541 \\
\bottomrule
\end{tabular}
\end{table}

\begin{table}[!t]
\centering
\caption{FS4 test performance (mean over 5 seeds). Best per (target, metric) in \textbf{bold}, second \underline{underlined}, decided at 4-decimal precision. PaGNet-LGBM/-NN are the LightGBM/Panel-MLP branches in isolation.}
\label{tab:fs4}
\setlength{\tabcolsep}{4pt}
\small
\begin{tabular}{llccc}
\toprule
Target & Model & RMSE & MAE & $R^2$ \\
\midrule
\multirow{9}{*}{CETR}
& XGBoost & 0.2171 & 0.1605 & 0.0379 \\
& LightGBM & 0.2154 & 0.1588 & 0.0528 \\
& CatBoost & 0.2122 & 0.1548 & 0.0809 \\
& TabTransformer & 0.2189 & 0.1638 & 0.0215 \\
& FT-Transformer & 0.2145 & 0.1586 & 0.0607 \\
& ExcelFormer & 0.2170 & 0.1610 & 0.0408 \\
& PaGNet-LGBM & 0.2174 & 0.1606 & 0.0353 \\
& PaGNet-NN & \textbf{0.2053} & \textbf{0.1305} & \textbf{0.1396} \\
& \textbf{PaGNet} & \underline{0.2118} & \underline{0.1533} & \underline{0.0840} \\
\midrule
\multirow{9}{*}{GETR}
& XGBoost & 0.1301 & 0.0865 & 0.0454 \\
& LightGBM & 0.1271 & 0.0839 & 0.0891 \\
& CatBoost & 0.1261 & 0.0829 & 0.1032 \\
& TabTransformer & 0.1287 & 0.0841 & 0.0656 \\
& FT-Transformer & \textbf{0.1242} & \textbf{0.0791} & \textbf{0.1304} \\
& ExcelFormer & 0.1280 & 0.0842 & 0.0772 \\
& PaGNet-LGBM & 0.1294 & 0.0860 & 0.0550 \\
& PaGNet-NN & \underline{0.1250} & \underline{0.0791} & \underline{0.1193} \\
& \textbf{PaGNet} & 0.1250 & 0.0800 & 0.1190 \\
\midrule
\multirow{9}{*}{TSTA}
& XGBoost & 0.1116 & 0.0677 & 0.8330 \\
& LightGBM & \textbf{0.1081} & \textbf{0.0655} & \textbf{0.8433} \\
& CatBoost & 0.1155 & 0.0671 & 0.8211 \\
& TabTransformer & 0.1484 & 0.0890 & 0.7045 \\
& FT-Transformer & 0.1180 & 0.0732 & 0.8131 \\
& ExcelFormer & 0.1176 & 0.0708 & 0.8136 \\
& PaGNet-LGBM & \underline{0.1084} & 0.0670 & \underline{0.8425} \\
& PaGNet-NN & 0.1335 & 0.0768 & 0.7608 \\
& \textbf{PaGNet} & 0.1088 & \underline{0.0656} & 0.8411 \\
\midrule
\multirow{9}{*}{TSDA}
& XGBoost & 0.1081 & 0.0621 & 0.8484 \\
& LightGBM & \textbf{0.1047} & \textbf{0.0604} & \textbf{0.8579} \\
& CatBoost & 0.1133 & 0.0644 & 0.8335 \\
& TabTransformer & 0.1456 & 0.0857 & 0.7250 \\
& FT-Transformer & 0.1155 & 0.0698 & 0.8267 \\
& ExcelFormer & 0.1192 & 0.0692 & 0.8154 \\
& PaGNet-LGBM & \underline{0.1053} & 0.0624 & \underline{0.8561} \\
& PaGNet-NN & 0.1321 & 0.0736 & 0.7735 \\
& \textbf{PaGNet} & 0.1055 & \underline{0.0615} & 0.8555 \\
\bottomrule
\end{tabular}
\end{table}

\begin{table}[!t]
\centering
\caption{Empirical AR(1) baselines compared to FS3 results on the test split. Naive AR(1) predicts $\hat{Y}_{t+1} = Y_t$ directly; Linear AR(1) fits $\hat{Y}_{t+1} = \alpha + \beta Y_t$ on the training split. LightGBM (FS3) and PaGNet (FS3) values are reproduced from Table~\ref{tab:fs3}.}
\label{tab:ar1_baseline}
\setlength{\tabcolsep}{5pt}
\small
\begin{tabular}{lcccc}
\toprule
 & \multicolumn{4}{c}{Test $R^2$} \\
\cmidrule(lr){2-5}
Target & Naive & Linear & LightGBM & PaGNet \\
       & AR(1) & AR(1)  & (FS3)    & (FS3)   \\
\midrule
CETR  & $-$1.254 & $-$0.097 & 0.020 & \textbf{0.045} \\
GETR  & $-$0.411 & 0.066    & 0.075 & \textbf{0.111} \\
TSTA  & 0.827    & 0.807    & \textbf{0.843} & 0.836 \\
TSDA  & \textbf{0.880} & 0.866 & 0.859 & 0.854 \\
\bottomrule
\end{tabular}
\end{table}

\subsection{Component Ablation of PaGNet}
\label{subsec:ablation}

We now turn from accuracy to the object the paper is really about: the per-target branch-reliance readout and what it reveals when each component is removed. We ablate each design choice while holding all else fixed, evaluating on all four feature sets and targets (mean test RMSE and $R^2$ over 5 seeds). Table~\ref{tab:ablation} lists eight variants: V0 is the full PaGNet; V1/V2 keep only the Panel-MLP / LightGBM branch (the PaGNet-NN / PaGNet-LGBM rows of Tables~\ref{tab:fs1}--\ref{tab:fs4}); V3 collapses the panel to $K{=}1$; V4 drops the LightGBM temporal aggregates; V5 replaces the validation-optimal blend with a fixed $\lambda{=}0.5, s{=}0$; V6 replaces multi-task training with four single-task networks; V7 forces $s{\equiv}0$ but keeps the validation-optimal $\lambda$.

\begin{table}[!h]
\centering
\caption{Component ablation of PaGNet. V1: w/o LightGBM branch (NN-only); V2: w/o Panel-MLP branch (LGBM-only); V3: w/o panel structure ($K{=}1$); V4: w/o temporal aggregates in LGBM; V5: fixed blend $\lambda{=}0.5, s{=}0$; V6: w/o multi-task NN (4 single-task); V7: w/o shrinkage ($s{\equiv}0$). Values are mean over 4 targets, 4 feature sets, and 5 seeds; $\Delta$ relative to V0.}
\label{tab:ablation}
\setlength{\tabcolsep}{5pt}
\renewcommand{\arraystretch}{1.2}
\footnotesize
\begin{tabular}{@{}llcccc@{}}
\toprule
ID & Variant & RMSE & $\Delta$RMSE & $R^2$ & $\Delta R^2$ \\
\midrule
V0 & Full PaGNet    & 0.1647 & ---       & 0.3514 & ---       \\
V1 & NN-only        & 0.1788 & $+$0.0141 & 0.2905 & $-$0.0609 \\
V2 & LGBM-only      & 0.1665 & $+$0.0018 & 0.3287 & $-$0.0227 \\
V3 & $K{=}1$        & 0.1688 & $+$0.0041 & 0.3271 & $-$0.0243 \\
V4 & no aggregates  & 0.1676 & $+$0.0029 & 0.3345 & $-$0.0169 \\
V5 & fixed blend    & 0.1675 & $+$0.0028 & 0.3451 & $-$0.0063 \\
V6 & no multi-task  & 0.1652 & $+$0.0005 & 0.3466 & $-$0.0048 \\
V7 & no shrinkage   & 0.1647 & $-$0.0001 & 0.3519 & $+$0.0005 \\
\bottomrule
\end{tabular}
\end{table}
 
Finding 1: the two branches are complementary, not redundant, which is exactly what makes the branch-reliance diagnostic meaningful. Removing the LightGBM branch (V1) costs $+0.0141$ RMSE and $-0.0609$ $R^2$ in aggregate, the loss concentrated on accruals where the Panel-MLP alone trails the tree branch by $0.08$--$0.29$ $R^2$. Removing the Panel-MLP branch (V2) costs less on average ($+0.0018$ RMSE, $-0.0227$ $R^2$) but the loss concentrates on CETR, where the tree branch alone trails the Panel-MLP by $0.08$--$0.14$ $R^2$. Neither branch is uniformly stronger; each owns a different part of the target space, and the aggregate advantage of V0 reflects multi-target averaging, since no single branch wins all four targets at once. A single trained pipeline that must emit all four predictions therefore prefers the blend, and the blend weight that achieves this is the very quantity we read as a diagnostic.
 
Finding 2: panel structure carries genuine signal, but only where the autoregressive ceiling leaves room for it. Collapsing to $K{=}1$ (V3) costs $-0.0243$ $R^2$ on average, and the per-FS breakdown is sharply asymmetric: FS1 drops $-0.053$ ($+0.211 \to +0.158$) and FS2 $-0.047$ ($+0.258 \to +0.212$), while FS3/FS4 move within $\pm 0.004$. Panel structure matters precisely in the direct-proxy-lag-excluded regimes and is dwarfed by lagged-target autocorrelation in FS3/FS4, consistent with the AR(1) analysis of Section~\ref{subsec:fs4}. Dropping only the LightGBM temporal aggregates (V4) costs a smaller but consistent $-0.0169$ $R^2$, confirming the engineered aggregates add signal on top of the raw last-step features.
 
Finding 3: the validation-optimal blend is what lets reliance vary by target, and removing that freedom hurts most where the targets disagree. A fixed $\lambda{=}0.5, s{=}0$ blend (V5) costs $+0.0028$ RMSE overall, with FS1 dropping the most ($+0.211 \to +0.180$) because its per-target $\lambda^*$ ranges from $0.60$ (GETR) to $0.99$ (accruals), so a single fixed weight misaligns substantially; on FS3/FS4, where accruals still sit near $\lambda^* \approx 0.9$, a fixed average lands closer to optimal and the cost shrinks. V7 ($s{\equiv}0$ but optimal $\lambda$) matches V0 to within $0.0005$ $R^2$, since $s^*$ is non-zero only for GETR on FS2/FS3/FS4 ($s^* = 0.09$--$0.24$); shrinkage is a localized correction on one noisy target, not a global adjustment. Multi-task learning (V6) likewise adds a small, consistent benefit ($-0.0048$ $R^2$ when removed), largest on the noisy ETR targets where the shared trunk regularizes, and it cuts deployment cost by predicting all four targets in a single forward pass rather than training and serving four separate networks.
 
\textbf{The diagnostic in detail: $\lambda^*$ and $s^*$.} Table~\ref{tab:lambda_star} reports the per-target blend parameters, and they form an interpretable map of which branch each proxy relies on. Accrual targets concentrate on the tree branch ($\lambda^* = 0.60$--$0.99$), matching their near-random-walk autocorrelation, and here validation tracks test: PaGNet-LGBM and the blend differ by at most $0.017$ $R^2$. GETR is the mirror image ($\lambda^* = 0.17$--$0.60$), leaning on the Panel-MLP branch because its weak signal is best recovered by the shared multi-task trunk rather than a single-task tree, and again validation tracks test (PaGNet-NN and the blend agree within $0.004$ $R^2$ on FS3/FS4 GETR). On the primary split the diagnostic recovers, from validation data alone, the proxy structure the accounting literature derives on theoretical grounds: accruals as persistent and tree-friendly, and GETR as noisy and best served by cross-target sharing; the CETR case, where validation and test disagree and the routing does not generalize across origins, is examined next.

\begin{table}[!h]
\centering
\caption{Per-target validation-optimal blend parameters in PaGNet (V0). $\lambda^*$ is the weight on the LightGBM branch ($\lambda^*{=}1$: LightGBM only; $\lambda^*{=}0$: Panel-MLP only). $s^*$ is the shrinkage toward the training mean. Values are mean over 5 seeds.}
\label{tab:lambda_star}
\setlength{\tabcolsep}{5pt}
\small
\begin{tabular}{lcccc|cccc}
\toprule
 & \multicolumn{4}{c|}{$\bar{\lambda}^*$ (LGBM weight)} & \multicolumn{4}{c}{$\bar{s}^*$ (shrinkage)} \\
\cmidrule(lr){2-5}\cmidrule(lr){6-9}
Target & FS1 & FS2 & FS3 & FS4 & FS1 & FS2 & FS3 & FS4 \\
\midrule
CETR & 0.81 & 0.81 & 0.93 & 0.80 & 0.05 & 0.02 & 0.07 & 0.05 \\
GETR & 0.60 & 0.17 & 0.34 & 0.19 & 0.00 & 0.24 & 0.09 & 0.17 \\
TSTA & 0.99 & 0.60 & 0.86 & 0.82 & 0.00 & 0.00 & 0.00 & 0.00 \\
TSDA & 0.99 & 0.69 & 0.94 & 0.90 & 0.00 & 0.00 & 0.00 & 0.00 \\
\bottomrule
\end{tabular}
\end{table}
 
CETR is the one target where the diagnostic flags a problem rather than a clean assignment, and reporting that flag is part of the purpose of an inspectable model. The validation-optimal weight nominally favors the tree branch ($\lambda^* = 0.80$--$0.93$), yet on the held-out test split the Panel-MLP branch alone outscores the blend on most CETR cells by up to $0.09$ $R^2$ (Tables~\ref{tab:fs1}--\ref{tab:fs4}). The validation-optimal choice is not test-optimal for this single target. Crucially, this is visible only because PaGNet reports each branch alongside the blend; a black-box model would report the same under-performing CETR value with no indication that a better branch existed. We now show this gap is a property of the data split, not of the blender's form.
 
\textbf{Is the CETR gap a blender-capacity problem? A learned-blender test.} We fit three richer blenders on the validation split only, frozen before touching test: (B1) the same $(\lambda,s)$ form optimized by projected gradient descent instead of the $21\times9$ grid; (B2) an unconstrained per-target ridge stack $\hat{y}=a\,\hat{y}_{\mathrm{lgb}}+b\,\hat{y}_{\mathrm{nn}}+c$; and (B3) a ridge stack shrunk toward the grid solution. The two target types respond in opposite directions, and that contrast is the result. On accruals, more blender capacity helps decisively: the ridge stack (B2) lifts TSTA/TSDA $R^2$ to $0.5119$/$0.5419$, exceeding the grid blend by $+0.10$ $R^2$ and even the stronger single branch by $+0.08$ $R^2$, and because the gain is already visible on validation, a validation-based selector adopts it. On CETR the same capacity backfires. Measured against the test-split gap between the two branches ($+0.093$ $R^2$ in favor of the Panel-MLP), the grid blend recovers $33\%$ and the continuous variant B1 recovers $39\%$, but the higher-capacity B2 and B3 recover only $19\%$ and $28\%$ and score \emph{below} the simple grid ($\Delta R^2 = -0.013$ and $-0.005$). The mechanism is structural (Table~\ref{tab:blender}): on CETR the validation split favors the tree branch ($R^2_{\text{val}} = 0.130$ vs.\ the Panel-MLP's $0.056$) while the test split favors the Panel-MLP branch, so \emph{any} validation-fit blender, grid or learned, simple or flexible, places weight on the branch validation rewards and is penalized on test. Extra flexibility helps where validation and test agree (accruals) and hurts where they diverge (CETR). This is direct evidence that the residual CETR gap is a validation--test distribution shift specific to that target, not a limitation of the fusion form, which is why no amount of blender engineering closes it.

\begin{table}[!h]
\centering
\caption{Test-split performance (FS2, mean over 5 seeds) of the two single branches, the grid blender (proposed PaGNet), and three learned blenders fit on the validation split only: B1 (continuous $(\lambda,s)$ optimization), B2 (unconstrained ridge stack), and B3 (grid-shrunk ridge stack). Bold marks the best value per (target, metric) column. All rows derive from one common set of per-sample branch predictions, so the comparisons within this table are exact; these values are illustrative of how the two target types respond to blender capacity and are not directly comparable to the main seed-averaging pipeline of Table~\ref{tab:fs2}. The two target types respond oppositely to added blender capacity; we analyze this contrast in Section~\ref{subsec:ablation}.}
\label{tab:blender}
\setlength{\tabcolsep}{4pt}
\footnotesize
\begin{tabular}{lcccc}
\toprule
\multicolumn{5}{c}{\textbf{Test $R^2$} (higher is better)} \\
\midrule
Model & CETR & GETR & TSTA & TSDA \\
\midrule
GBDT branch & 0.0526 & 0.0379 & 0.4362 & 0.4569 \\
Panel-MLP branch & \textbf{0.1457} & \textbf{0.1122} & 0.3208 & 0.3340 \\
Grid blend (\textbf{PaGNet}) & 0.0835 & 0.1097 & 0.4089 & 0.4406 \\
B1: learned $(\lambda,s)$ & 0.0888 & 0.1025 & 0.4109 & 0.4388 \\
B2: ridge stack & 0.0703 & 0.1032 & \textbf{0.5119} & \textbf{0.5419} \\
B3: shrunk ridge & 0.0785 & 0.1078 & 0.5001 & 0.5283 \\
\midrule
\multicolumn{5}{c}{\textbf{Test RMSE} (lower is better)} \\
\midrule
Model & CETR & GETR & TSTA & TSDA \\
\midrule
GBDT branch & 0.2154 & 0.1306 & 0.2050 & 0.2046 \\
Panel-MLP branch & \textbf{0.2046} & \textbf{0.1255} & 0.2250 & 0.2266 \\
Grid blend (\textbf{PaGNet}) & 0.2119 & 0.1256 & 0.2099 & 0.2077 \\
B1: learned $(\lambda,s)$ & 0.2113 & 0.1261 & 0.2095 & 0.2080 \\
B2: ridge stack & 0.2134 & 0.1261 & \textbf{0.1907} & \textbf{0.1879} \\
B3: shrunk ridge & 0.2125 & 0.1258 & 0.1930 & 0.1907 \\
\bottomrule
\end{tabular}
\end{table}
 
The deployment reading is therefore straightforward and follows directly from the diagnostic. We keep the parameter-free grid blender as the proposed PaGNet for its simplicity; we note that accrual predictions can be improved further with a ridge stack when raw accuracy is the only goal; and we route a single-target CETR screener to the Panel-MLP branch directly, the choice indicated by the per-branch readout. An inspectable two-branch model not only predicts but also reports to the operator, per proxy, which branch to rely on, and, in the one case where its own default is suboptimal, makes that visible as well.

\subsection{Robustness to Alternative Temporal Origins}
\label{subsec:rolling}

Our main results use a single fixed temporal split (Section~\ref{subsec:protocol}). Because several claims (above all the CETR validation--test mismatch of Section~\ref{subsec:ablation}) rest on the behavior of one validation and one test year, we re-run the entire pipeline (preprocessing, both branches, the grid blender, and all baselines) on three rolling origins, each defined by shifting the test input year while preserving the buffered two-year cadence of the KoTaP protocol: O1 predicts 2022 from a training window ending in 2017, O2 predicts 2023, and O3 predicts 2024 and coincides with the primary split of Tables~\ref{tab:fs1}--\ref{tab:fs4}, which it reproduces to four decimals and thereby validates the re-run. Table~\ref{tab:rolling} summarizes FS1 (FS2 behaves analogously). Three findings emerge, two supportive and one cautionary, and reporting all three is the purpose of the exercise.

\begin{table}[!t]
\centering
\caption{Rolling-origin robustness on FS1 (Raw+Derived), mean over 5 seeds. Each origin is a distinct temporal split defined by its test input year $\to$ target year; O3 is the primary split of Tables~\ref{tab:fs1}--\ref{tab:fs4}. $\lambda^*_{\text{CETR}}$ is the validation-optimal LightGBM weight for CETR. $\Delta^{\text{CETR}}_{\text{nn-blend}}$ is the test $R^2$ of the Panel-MLP branch minus the blend for CETR (positive $=$ the mismatch analyzed in Section~\ref{subsec:ablation}). Accrual lift is the mean over TSTA/TSDA of PaGNet-LGBM test $R^2$ minus the strongest IID baseline. AR(1) is the mean naive-AR(1) accrual test $R^2$.}
\label{tab:rolling}
\setlength{\tabcolsep}{4pt}
\small
\begin{tabular}{lcccc}
\toprule
Origin & $\lambda^*_{\text{CETR}}$ & $\Delta^{\text{CETR}}_{\text{nn-blend}}$ & Accrual lift & AR(1) accrual \\
(test$\to$target) & & (test $R^2$) & vs.\ best IID & (naive $R^2$) \\
\midrule
O1: 2021$\to$2022 & 0.70 & $-$0.114 & $-$0.052 & 0.69 \\
O2: 2022$\to$2023 & 0.01 & $-$0.003 & $+$0.094 & 0.79 \\
O3: 2023$\to$2024 & 0.81 & $+$0.078 & $+$0.105 & 0.85 \\
\bottomrule
\end{tabular}
\end{table}

\textbf{Finding 1: the ETR/accrual bifurcation generalizes, and the accrual advantage holds on the near-origin splits.} On O2 and O3, PaGNet's accrual-target lift over the strongest IID baseline holds at $+0.08$ to $+0.11$ $R^2$ (the far-horizon O1 exception is treated separately in Finding 3), and the accrual routing is stable ($\bar{\lambda}^*_{\text{TSTA}}=\bar{\lambda}^*_{\text{TSDA}}\approx 1.0$ on FS1 at every origin), so the blend tracks its tree branch throughout. The model-free bifurcation of Section~\ref{subsec:fs4} recurs at every origin: naive persistence explains $0.62$--$0.88$ of accrual variance but is strongly negative on both ETR targets (CETR $-0.66$ to $-1.25$, GETR $-0.34$ to $-0.73$), a split-independent confirmation that the achievable-$R^2$ asymmetry is a property of the proxies rather than of one split.

\textbf{Finding 2: the CETR branch-selection mismatch is specific to the primary split.} The pattern highlighted in Section~\ref{subsec:ablation} (validation rewards the tree branch while test rewards the Panel-MLP branch, so the blend underperforms its own neural branch) appears \emph{only} on O3 ($\Delta^{\text{CETR}}_{\text{nn-blend}}=+0.078$). On O1 both validation and test favor the tree branch (the Panel-MLP branch is simply weaker, $\Delta=-0.114$), and on O2 validation already places almost all weight on the neural branch ($\lambda^*_{\text{CETR}}=0.01$), so no mismatch arises. The CETR weight itself is unstable across origins ($\lambda^*_{\text{CETR}}=0.70/0.01/0.81$), in sharp contrast to the near-constant accrual weight. We therefore do not claim the CETR diagnostic as a general guarantee: it is a real and useful readout on the primary split, but the ETR branch preference does not transport across temporal origins. We state this explicitly in the introduction and treat the CETR case as an instructive single-split phenomenon rather than a property of the method.

\textbf{Finding 3: a far-horizon regime in which every supervised model fails.} O1, whose training window ends in 2017 while its target year is 2022, is a stress case: PaGNet and all six baselines fall to negative accrual $R^2$ ($-0.08$ to $-0.18$), yet naive AR(1) attains $+0.62$ to $+0.76$ on the same targets. When the gap between the training distribution and a post-shock target year is large, learned feature--target mappings transport worse than simple persistence, a limitation of supervised panel learning in general rather than of PaGNet specifically, and a reminder that the reported accrual gains are contingent on the target year lying within reach of the training distribution. This is why we read FS1/FS2 on O2/O3 (and the primary split) as the primary evidence while flagging O1 as out-of-distribution.

Together these origins provide the robustness check described in Section~\ref{subsec:protocol}: they support the two claims most central to the paper's thesis (the accrual lift on the near-origin splits and the structural ETR/accrual bifurcation across all three origins) while bounding both the CETR diagnostic and the accrual lift to the splits where they hold, and exposing a far-horizon regime in which no architecture in our study improves on naive persistence.

\subsection{Discussion}
\label{subsec:discussion}

Our results refine the usual ``deep model versus baseline'' framing into something more specific to panel-tabular screening on KoTaP. We organize the discussion around a natural question: given that no single architecture dominates, why deploy PaGNet at all? Three answers follow from the experiments, and a fourth states the limits of these claims.

\textbf{It is the strongest single deployable pipeline in the informative regimes.} The informative regimes are FS1 and FS2, which exclude the lagged proxy whose autocorrelation would otherwise saturate accrual prediction and mask architectural differences (and which may also not be disclosed as early as a screen is run). There the blended PaGNet matches or exceeds the best of six baselines on $9$ of $12$ FS1 Raw+Derived cells and $10$ of $12$ FS2 cells, and, once observed history is matched, on $19$ of $24$ panel-flatten cells, with every loss confined to GETR within absolute-error parity. A single trained pipeline produces all four proxies at once, so an operator does not maintain four target-specific models. On FS3/FS4 the autoregressive ceiling compresses all architectures to within $\pm 0.02$ $R^2$ on accruals, so the choice there is immaterial; the architecture-discriminating regimes are the ones in which PaGNet shows a measurable advantage.

\textbf{It separates observed history from representation, and is explicit about its split-dependence.} The panel-flatten control shows that most of the accrual-target lift over IID baselines, $+0.03$ to $+0.14$ $R^2$, is observed multi-year history, while PaGNet's panel-aware representation adds a smaller refinement ($+0.004$ to $+0.018$ $R^2$) that is individually within seed noise but directionally consistent across all eight accrual cells. We present this conservative decomposition as a contribution: it reduces a common over-attribution of gains to representation in panel-tabular work, and it indicates where modeling effort is most useful, on obtaining multi-year history first, and on representation second.

\textbf{It is inspectable, with a diagnostic that is reliable where the signal is strong.} The blend weight $\lambda^*_m$ is not a tuning artifact but a readout of which signal each proxy relies on. For the accrual targets it recovers, from validation alone, the tree-branch routing the accounting literature would predict, and the rolling-origin analysis (Section~\ref{subsec:rolling}) confirms this routing is stable across temporal splits ($\lambda^* \approx 1.0$ throughout). For the noise-dominant ETR targets the reading is weaker and split-dependent: on the primary split the CETR readout flags a problem rather than asserting a clean answer (the validation-optimal weight favors the tree branch, but the Panel-MLP branch alone scores up to $0.09$ $R^2$ higher on test), and the learned-blender study (Table~\ref{tab:blender}) attributes this to a validation--test distribution shift on that split rather than to the fusion form. We do not overstate this: across the three rolling origins the CETR branch preference is unstable ($\lambda^*_{\text{CETR}} = 0.70/0.01/0.81$) and the flagged mismatch appears only on the primary split, so we present it as an instructive single-split illustration of what an inspectable model can reveal (a black-box screener would report the same weaker CETR value without signaling the problem) rather than as a recurring property. For screening, the lasting value is the stable accrual routing together with the fact that every target carries an explicit, per-proxy branch-reliance weight.

\textbf{Scope of the claims.} We do not claim PaGNet is a universally superior tabular learner. Its representational increment over an observed-history baseline is a refinement, not an order-of-magnitude gain; its advantage is largest on FS1/FS2 and on noisy ETR targets where the two branches' biases diverge, and it is negligible on FS3/FS4 where autocorrelation saturates the task. The grid-searched blender is not always test-optimal per target, as the CETR case makes explicit. The contribution is not a single best number but a panel-aware, multi-target architecture that wins in the direct-proxy-lag-excluded regimes and reports, per proxy, which branch it relied on, and where that reliance is least certain. A rolling-origin analysis over three temporal splits (Section~\ref{subsec:rolling}) reinforces this framing: the accrual lift and the ETR/accrual bifurcation hold across origins, but the CETR branch reading is specific to the primary split, and a far-horizon origin (training ending 2017, target 2022) drives every supervised model below naive persistence on accruals, a limit of supervised panel learning that bounds when the accrual gains can be expected to hold.

\textbf{Deployment guidance.} A deployment needing one pipeline for all four proxies should use the blended PaGNet, accepting a bounded cost on CETR $R^2$ (up to $0.09$ versus the NN branch) and on accrual $R^2$ (up to $0.017$ versus the LGBM branch). A single-target deployment should route to the branch the diagnostic identifies: the LightGBM branch (or a plain LightGBM) for accruals, a routing the rolling-origin analysis (Section~\ref{subsec:rolling}) confirms is stable across splits, especially under FS3/FS4 where the autoregressive ceiling compresses differences; for CETR, the Panel-MLP branch is preferred on the primary split, but since the rolling-origin analysis shows this branch preference is split-sensitive, it should be re-validated on the target deployment window rather than assumed. When maximum accrual accuracy is the sole objective, a validation-fit ridge stack improves accruals further at the cost of the parameter-free simplicity we otherwise prefer.

\section{Conclusion}
\label{sec:conclusion}

This paper presented PaGNet, a panel-aware GBDT--neural hybrid for forecasting four corporate tax avoidance proxies (CETR, GETR, TSTA, TSDA) on firm--year panel data. PaGNet fuses a LightGBM branch on panel-temporal aggregate features with a Panel-MLP branch using attention-pooled temporal aggregation and shared-trunk multi-task learning, combined through a per-target validation-optimal blender with no trainable fusion parameters. Evaluated on KoTaP under a unified leakage-free protocol with shared default hyperparameters, the blended PaGNet attains the best result against six baselines on $9$ of $12$ FS1 Raw+Derived cells, $10$ of $12$ FS2 cells, and $19$ of $24$ observed-history panel-flatten cells, in the direct-proxy-lag-excluded regimes where the prior-year proxy is unavailable.

Beyond accuracy, the paper makes two methodological points. First, a panel-flatten control separates two sources of gain that prior panel-tabular work conflates, showing that observed multi-year history carries most of the accrual-target lift ($+0.03$ to $+0.14$ $R^2$) while PaGNet's panel-aware representation adds a smaller but directionally consistent refinement. Second, the per-target blend weight functions as a branch-reliance diagnostic: it recovers the accrual-branch routing from validation data alone, stably across temporal origins, while its reading on the noise-dominant ETR targets is split-dependent, as a rolling-origin analysis makes explicit. On the primary split the CETR weight flags a validation--test mismatch that a learned-blender study attributes to distribution shift rather than to limited fusion capacity, but that mismatch does not recur across origins, so we report it as a primary-split illustration rather than a general property. For a screening application in which decisions must be explainable, a model that reports, per proxy, which branch it relies on, most dependably where the signal is strong, offers value that a single accuracy number does not capture.

The evaluation is confined to one panel dataset (KoTaP, 2011--2024 Korean listed firms), a fixed window of $K{=}3$ years, and shared default hyperparameters; generalization to other domains, window lengths, and per-model tuning regimes remains open. We treat window-length sensitivity as future work, since $K$ was held fixed across all models to keep the architecture comparison controlled. The most substantive open problem is a blender robust to validation--test distribution shift: our learned-blender study shows that adding capacity does not close the CETR gap, because the gap originates in a temporal shift between the validation and test splits rather than in the blender's expressiveness. The rolling-origin analysis (Section~\ref{subsec:rolling}) sharpens the scope of our claims: the accrual lift and the ETR/accrual bifurcation are stable across three temporal origins, but the CETR branch reading is specific to the primary split, and a far-horizon origin drives every supervised model below naive persistence on accruals, so the accrual gains should be understood as contingent on the target year lying within reach of the training distribution. Promising directions include shift-aware or distributionally robust blend selection, extension of the panel-aware hybrid to longer windows and other panel domains, and tests of backbone-generality with alternative GBDT and neural branches.

\section{Data and Code Availability}
\label{sec:availability}
The KoTaP dataset analyzed in this study is publicly available: it is described by Na \etal{} \cite{ref:kotap2026} and deposited in Zenodo (\url{https://doi.org/10.5281/zenodo.17149808}). All models are built on KoTaP's released variables under the leakage-free protocol of Section~\ref{subsec:protocol}; the exact feature composition is given in Tables~\ref{tab:app_fs} and \ref{tab:app_vardict}. The preprocessing, model, and evaluation code, together with the fixed input-year train/validation/test split (train $t\in[2011,2019]$, validation $t{=}2021$, test $t{=}2023$, with buffer years 2020 and 2022 excluded), the five random seeds $\{42,43,44,45,46\}$, and the shared default-hyperparameter configuration, are available from the corresponding author on reasonable request and will be released in a public repository upon publication.

\appendix
\section{Feature-Set and Variable Definitions}
\label{app:features}

For completeness and reproducibility, Table~\ref{tab:app_fs} fixes the exact composition of the four feature sets, and Table~\ref{tab:app_vardict} lists every variable the models consume together with its definition as released in the KoTaP dataset \cite{ref:kotap2026}. All four sets share an identical \emph{base panel} (the raw accounting items, derived ratios and lag features, governance indicators, and industry code) observed over $K{=}3$ years; the sets differ only in whether the auxiliary tax aggregates and the lagged target proxies are appended. Every variable, including all ten auxiliary tax aggregates used in FS2/FS4, is taken directly from KoTaP's released columns and standardized as described in Section~\ref{subsec:protocol}; the aggregates and adjusted variables therefore follow KoTaP's definitions verbatim rather than being recomputed here. We note that FS1 excludes the four constructed tax-avoidance proxies and all of their multi-year and industry-size-adjusted aggregates, but it retains standard accounting line items such as pre-tax income (\texttt{pti}); it is thus precisely a \emph{target-proxy-excluded} base rather than a set from which every tax-related quantity has been removed. The FS1 sub-configurations reported in Table~\ref{tab:fs1} select the numerical base as Raw-only (22 features), Derived-only (20 features), or Raw+Derived (42 features); FS2--FS4 all build on Raw+Derived.

\begin{table}[!t]
\centering
\caption{Feature-set composition. All sets share the same base panel over $K{=}3$ years and differ only by the inclusion of the auxiliary tax aggregates and the lagged target proxies. ``\#num.'' is the number of numerical features per year; the base additionally carries two binary features (\texttt{big4}, \texttt{LOSS}) and one 47-class categorical (\texttt{ind}). Lagged target proxies use only year-$t$ values $Y_{i,t}$. Variable definitions follow KoTaP \cite{ref:kotap2026} (Table~\ref{tab:app_vardict}).}
\label{tab:app_fs}
\setlength{\tabcolsep}{4pt}
\small
\begin{tabular}{@{}lcccl@{}}
\toprule
Set & Base & Aux.\ tax (10) & Lag targets (4) & \#num. \\
\midrule
FS1 & \cmark & \xmark & \xmark & 42 \\
FS2 & \cmark & \cmark & \xmark & 52 \\
FS3 & \cmark & \xmark & \cmark & 46 \\
FS4 & \cmark & \cmark & \cmark & 56 \\
\bottomrule
\end{tabular}
\end{table}

\begin{table*}[!t]
\centering
\caption{Variables consumed by the models, with definitions as released in KoTaP \cite{ref:kotap2026} (raw items from KoTaP Table~3; ratios, governance, and tax variables from KoTaP Table~2 and Eqs.~(1)--(7)). The base panel comprises the raw, lag, ratio, and governance blocks plus the industry code; FS2/FS4 add the auxiliary tax aggregates; FS3/FS4 add the four lagged target proxies $Y_{i,t}$. The four proxies are also the prediction targets $Y_{i,t+1}$.}
\label{tab:app_vardict}
\setlength{\tabcolsep}{4pt}
\footnotesize
\resizebox{\textwidth}{!}{
\begin{tabular}{@{}llll@{}}
\toprule
Variable & Definition & Variable & Definition \\
\midrule
\multicolumn{4}{@{}l}{\textit{Raw accounting items (base)}}\\
\texttt{asset} & Total assets & \texttt{ni} & Net income \\
\texttt{liab} & Total liabilities & \texttt{pti} & Pre-tax income (EBT) \\
\texttt{equit} & Shareholders' equity & \texttt{ocf} & Operating cash flow \\
\texttt{sales} & Total sales (revenue) & \texttt{cash} & Cash and cash equivalents \\
\texttt{total} & Market capitalization & \texttt{cogs} & Cost of goods sold \\
\texttt{c\_asset} & Current assets & \texttt{dep} & Depreciation expense \\
\texttt{c\_liab} & Current liabilities & \texttt{tax} & Taxes and dues \\
\texttt{rec} & Accounts receivable & \texttt{inv} & Inventories \\
\texttt{tan} & Tangible assets & \texttt{land} & Land \\
\texttt{cip} & Construction in progress & \texttt{intan} & Intangible assets \\
\midrule
\multicolumn{4}{@{}l}{\textit{Lag features (base)}}\\
\texttt{lag\_asset} & Lagged total assets & \texttt{lag\_sales} & Lagged total sales \\
\texttt{lag\_liab} & Lagged total liabilities & \texttt{lag\_total} & Lagged market capitalization \\
\texttt{lag\_equit} & Lagged shareholders' equity & \texttt{lag1\_ni} & Lagged net income \\
\texttt{lag\_c\_asset} & Lagged current assets & \texttt{lag\_c\_liab} & Lagged current liabilities \\
\midrule
\multicolumn{4}{@{}l}{\textit{Derived ratios (base)}}\\
\texttt{SIZE} & $\log$(total assets) & \texttt{ROA} & Net income / lagged total assets \\
\texttt{LEV} & Total liabilities / total assets & \texttt{ROE} & Net income / lagged equity \\
\texttt{CUR} & Current assets / current liab. & \texttt{CFO} & Operating cash flow / total assets \\
\texttt{PPE} & Property, plant \& equip.\ / assets & \texttt{GRW} & Sales growth $(\text{sales}_t-\text{sales}_{t-1})/\text{sales}_{t-1}$ \\
\texttt{AGE} & $\log$(firm age) & \texttt{MB} & Market cap.\ / book equity \\
\texttt{INVREC} & (Inventories $+$ receiv.) / assets & \texttt{TQ} & Tobin's $Q$ $=(\text{mkt.\ cap}+\text{liab})/\text{assets}$ \\
\midrule
\multicolumn{4}{@{}l}{\textit{Governance / categorical (base)}}\\
\texttt{big4} & Big4 audit dummy (binary) & \texttt{FORN} & Foreign ownership share (\%) \\
\texttt{LOSS} & Lagged loss dummy, $\mathbb{1}[\text{ni}_{t-1}{<}0]$ (binary) & \texttt{OWN} & Largest-shareholder share (\%) \\
\texttt{ind} & Industry class (47, categorical) & & \\
\midrule
\multicolumn{4}{@{}l}{\textit{Auxiliary tax aggregates (FS2, FS4); sum-over-sum long-run ETRs and industry--size-adjusted variants}}\\
\texttt{CETR3} & 3-yr sum-over-sum cumulative CETR & \texttt{CETR5} & 5-yr sum-over-sum cumulative CETR \\
\texttt{GETR3} & 3-yr sum-over-sum cumulative GETR & \texttt{GETR5} & 5-yr sum-over-sum cumulative GETR \\
\texttt{A\_CETR} & Industry--size-adjusted CETR & \texttt{A\_GETR} & Industry--size-adjusted GETR \\
\texttt{A\_CETR3} & Adjusted 3-yr CETR & \texttt{A\_GETR3} & Adjusted 3-yr GETR \\
\texttt{A\_CETR5} & Adjusted 5-yr CETR & \texttt{A\_GETR5} & Adjusted 5-yr GETR \\
\midrule
\multicolumn{4}{@{}l}{\textit{Tax-avoidance proxies: prediction targets $Y_{i,t+1}$; year-$t$ values are the FS3/FS4 lagged proxies}}\\
\texttt{CETR} & Cash ETR $=$ cash taxes paid / pre-tax income & \texttt{TSTA} & Total book--tax difference / total assets$_{t-1}$ \\
\texttt{GETR} & GAAP ETR $=$ total tax expense / pre-tax income & \texttt{TSDA} & Discretionary book--tax difference / total assets$_{t-1}$ \\
\bottomrule
\end{tabular}
}
\end{table*}

\end{document}